%% file: main.tex
\documentclass[twoside,11pt]{article}
\usepackage{jair, theapa, rawfonts}

\firstpageno{1}

\usepackage{multirow}
\usepackage{vector}
\usepackage{amsmath,amssymb,amsfonts}
\usepackage{xcolor}
\usepackage{siunitx}
\usepackage{booktabs}
\usepackage{subcaption}
\usepackage{ulem}
\newcommand{\ra}[1]{\renewcommand{\arraystretch}{#1}}

\newcolumntype{R}{>{$}r<{$}}

\usepackage{pgfplots}
\pgfplotsset{compat=newest}
\usepgfplotslibrary{groupplots}
\usepgfplotslibrary{dateplot}

\newcommand{\citet}[1]{\citeauthor{#1} \citeyear{#1}}
\usepackage{graphicx}
\usepackage{url}
\usepackage[capitalize]{cleveref}

\begin{document}

\title{Conditional Approximate Normalizing Flows for Joint Multi-Step Probabilistic Forecasting with Application to Electricity Demand}
\author{\name Arec Jamgochian$^1$ \email
        arec@stanford.edu \\
       \name Di Wu$^2$ \email di.wu5@mail.mcgill.ca \\
       \name Kunal Menda$^1$ \email kmenda@stanford.edu \\
       \name Soyeon Jung$^1$ \email soyeonj@stanford.edu \\
       \name Mykel J. Kochenderfer$^1$ \email mykel@stanford.edu \\
       \addr $^1$Stanford University, Stanford, CA 94305 USA \\
       \addr $^2$McGill University, Montreal, Quebec H3A 0G4 Canada \\
       }

% For research notes, remove the comment character in the line below.
% \researchnote

\maketitle

\begin{abstract}
\input{0-abstract}
\end{abstract}

\input{1-intro}
\input{2-motivation}
\input{3-method}

\input{4-experiments}

\input{5-conclusion}
\section*{Acknowledgements}
This material is based upon work supported by the National Science Foundation Graduate Research Fellowship Program under Grant No. DGE-1656518. Any opinions, findings, and conclusions or recommendations expressed in this material are those of the author(s) and do not necessarily reflect the views of the National Science Foundation. 
This work is also supported by the COMET K2---Competence Centers for Excellent Technologies Programme of the Federal Ministry for Transport, Innovation and Technology (bmvit), the Federal Ministry for Digital, Business and Enterprise (bmdw), the Austrian Research Promotion Agency (FFG), the Province of Styria, and the Styrian Business Promotion Agency (SFG). 

\vskip 0.2in
\bibliography{references}
\bibliographystyle{theapa}

\end{document}

%% file: 0-abstract.tex
Some real-world decision-making problems require making probabilistic forecasts over multiple steps at once. 
However, methods for probabilistic forecasting may fail to capture correlations in the underlying time-series that exist over long time horizons as errors accumulate. 
One such application is with resource scheduling under uncertainty in a grid environment, which requires forecasting electricity demand that is inherently noisy, but often cyclic. 
In this paper, we introduce the \textit{conditional approximate normalizing flow} (CANF) to make probabilistic multi-step time-series forecasts when correlations are present over long time horizons. 
We first demonstrate our method's efficacy on estimating the density of a toy distribution, finding that CANF improves the KL divergence by one-third compared to that of a Gaussian mixture model while still being amenable to explicit conditioning. 
We then use a publicly available household electricity consumption dataset to showcase the effectiveness of CANF on joint probabilistic multi-step forecasting. 
Empirical results show that conditional approximate normalizing flows outperform other methods in terms of multi-step forecast accuracy and lead to up to 10x better scheduling decisions. 
Our implementation is available at \url{https://github.com/sisl/JointDemandForecasting}.

%% file: 1-intro.tex
\section{Introduction} \label{sec:intro}
Electric load forecasting is one of the core problems for modern power grids. 
Depending on the forecasting horizon, there are three types of load forecasts: short-term (minutes or hours ahead), mid-term (months ahead), and long-term (a few years ahead). 
Short-term load forecasting (STLF), which is the main focus of this paper, is typically used to assist real-time energy dispatching. 
Accurate electric load forecasting is important for the safe and efficient operation of modern power grids. A medium-sized utility company can save millions of dollars annually by improving the accuracy of forecasts~\cite{tribble2003relationship}.

Short-term load forecasting can be very difficult, especially with the increasing uncertainties in both power consumption and power generation.
For example, renewable energy generation, which has grown exponentially in the last ten years~\cite{murdock2019renewables}, is highly dependent on weather conditions. 
With the increasing number of adopted appliances, the total power consumption pattern also becomes more complex. 

Demand forecasting also has many potential applications on the customer side. With the growing popularity of home batteries and solar panels, demand forecasting can help guide sustainable battery transport and storage. With the increasing adoption of electric vehicles (EV), EV charging scheduling is important for the safe operation of power grids. Accurate demand forecasting will be essential in guiding charge scheduling. In general, planning can be improved by incorporating uncertainty into forecasts. 

The field of probabilistic load forecasting (PLF) seeks to predict a probability distribution of future demand. 
Specifically, methods aim to characterize $p(y_{t+K} \mid \vect{y}_{t-L:t})$, where $y_i$ is the load at time $i$, $L$ is a lag index, and $K$ is the index for future load.
By capturing variability in future loads, planning using PLF can be much more robust.
However, one significant drawback to the methods typically used for probabilistic short-term load forecasting is that they often only consider forecasting a single future time-step, i.e. where $K=1$~\cite{hong2016probabilistic}.

While single-step forecasts are useful for next-step planning, the dynamic allocation of resources often requires considering longer time horizons. For example, to schedule the charging of a set of electric vehicles when external demand is at a minimum, one would need to forecast a distribution over trajectories of possible future demand. With previous work in PLF, building models for multiple-step forecasting generally requires applying one of two methods:
\begin{enumerate}
    \item \textit{Independent}: build separate models for different future times (i.e. $p(y_{t+1} \mid \vect{y}_{t-L:t})$, $p(y_{t+2} \mid \vect{y}_{t-L:t})$, etc.)
    \item \textit{Iterative}: repeatedly append samples from a single-step model as inputs to sample further into the future (i.e. $\hat{y}_{t+1} \sim p(y_{t+1} \mid \vect{y}_{t-L:t})$, $\hat{y}_{t+2} \sim p(y_{t+2} \mid \vect{y}_{t-L+1:t}, \hat{y}_{t+1})$, etc.) 
\end{enumerate}

Forecasting using independent models cannot capture correlations between different time-steps. When sampling independent models, we are often left with noisy, discontinuous electricity demand trajectories. Iterative methods generate smooth trajectories, but errors can accumulate with every step, which is especially problematic if time-series exhibit correlations over long time horizons.
 
An alternative to iterative and independent methods is one in which we explicitly model the joint distribution $p(\vect{y}_{t+1:t+K}\mid \vect{y}_{t-L:t})$. 
This paper briefly discusses extensions of frequently used STLF methods to do joint multi-step demand forecasting. We then introduce an approach called the \textit{conditional approximate normalizing flow} (CANF). To the best of our knowledge, this is the first attempt at conditioning a normalizing flow to do multi-step forecasting.

We demonstrate the ability of an approximate normalizing flow to estimate the joint density of a toy distribution better than a Gaussian Mixture Model (GMM) while still being amenable to explicit conditioning.
The forecasting performance of the CANF is then compared to other joint and iterative multi-step probabilistic forecasting methods. Additionally, we use task scheduling under uncertainty to compare decision quality when using different forecasting methods. We observe that the CANF achieves consistently superior performance in both predictive accuracy and decision quality. In summary, the main contributions of this paper are to:
\begin{itemize}
    \item Introduce the conditional approximate normalizing flow to perform explicit joint multi-step probabilistic forecasting, 
    \item Demonstrate the efficacy of using approximate normalizing flows to estimate probability densities while remaining amenable to explicit conditioning, and,
    \item Empirically compare the forecasting accuracy and resulting decision quality of the CANF when predicting household electricity consumption.
\end{itemize}

The electricity demand forecasting domain has the properties that a) strong correlations exist over long time horizons that are not easily captured by iterative methods, and b) there is not enough data for a GMM to express a joint model sufficiently well. We believe our method would perform well on other domains with these same characteristics. 

The remainder of this paper is organized as follows: \cref{sec:back} outlines motivation and related work, \cref{sec:method} describes methods for joint multi-step forecasting and how to use those forecasts for scheduling under uncertainty, \cref{sec:experiments} highlights experiments and results, and \cref{sec:discussion} concludes.

%% file: 2-motivation.tex
\section{Motivation and Related Work} \label{sec:back}
This section motivates the need for accurate multi-step probabilistic load forecasting and discusses relevant literature on the topic.

\subsection{Using forecasts for planning}

\begin{figure}[htbp]
\centerline{\includegraphics[trim=0 150 0 0,clip,width = 3.2in]{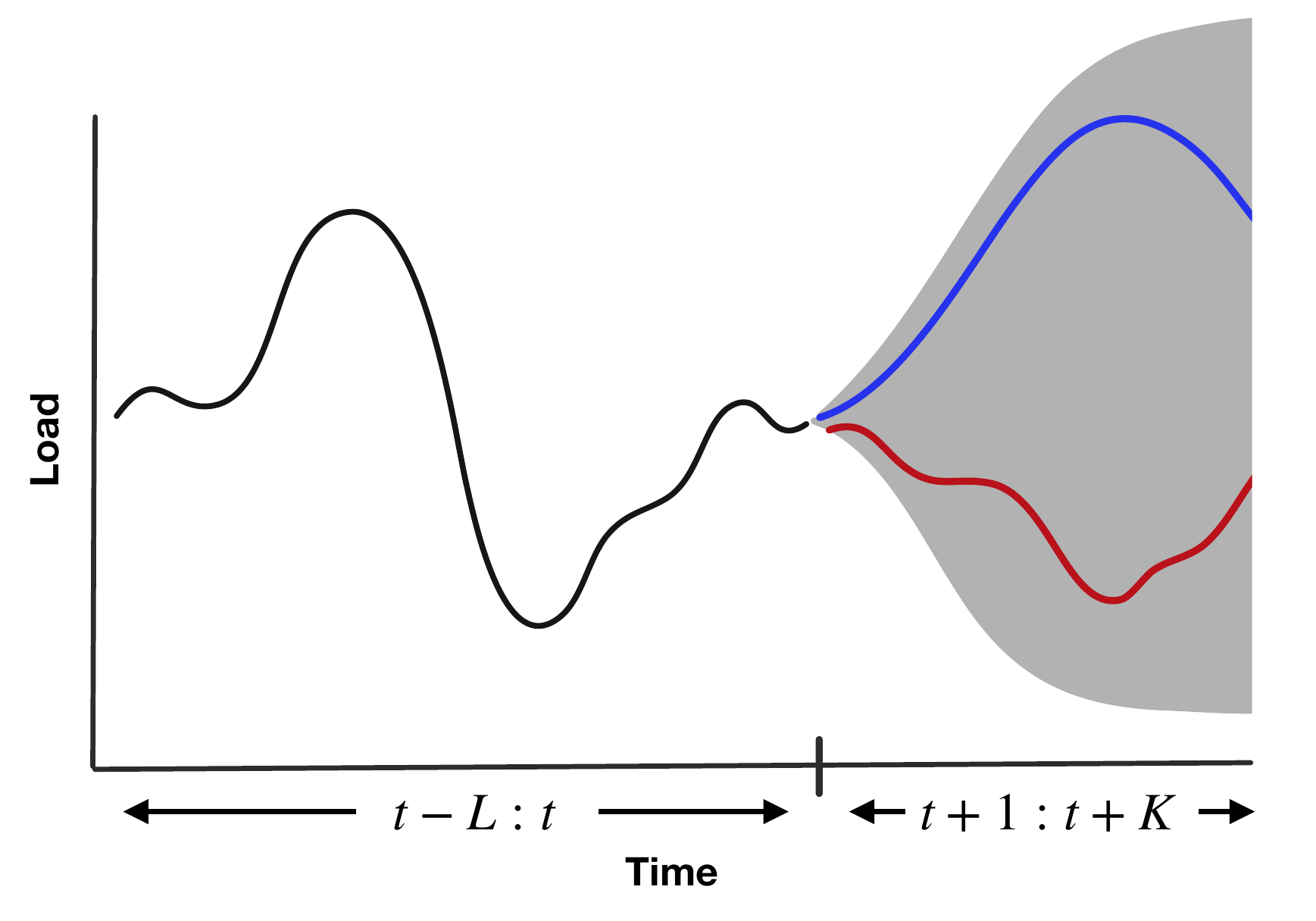}}
\caption{Making robust scheduling decisions in a grid environment requires quantifying uncertainty over multiple steps. Different possible demand trajectories (red and blue) can inform whether to act now or later.}
\label{fig:decisions}
\end{figure}

To motivate the need for multi-step forecasting, consider the following example, as illustrated in~\cref{fig:decisions}: we are given a fixed time-window to charge an electric vehicle. For sustainability (and potentially cost) considerations, we want to charge when external electricity demand is at a minimum.  At each time step, we must forecast external electricity demand at future points in time, and decide whether waiting for potentially lower external demand is worth the risk. Our decision will inevitably be based on our confidence that external demand will decrease and, if so, by how much. Without scheduling, EV charging can cause extra energy costs for customers and cause extremely high peak power consumption, which is very dangerous for the power grid infrastructures~\cite{wu2017two}. Similarly, many planning problems regarding the allocation of resources in a power grid environment rely on multi-step forecast `trajectories'. 

Dynamic energy management \cite{wytock2017dynamic,moehle2019dynamic} is a form of robust model predictive control that makes single-step decisions based on long-term performance against many sampled energy scenarios, thus considering future uncertainty. Planning with Monte Carlo Tree Search \cite{couetoux2011continuous,browne2012survey} requires sampling many futures while searching the action space to make higher-reward decisions. 

Explicit multi-step forecast distributions concisely characterize `risk' and may sometimes be used for scalable, risk-sensitive planning through convex optimization. One such famous example is portfolio optimization \cite{markovitz1959portfolio}, in which the explicitly forecasted covariance of asset returns is used to make risk-bounded portfolio allocations. 
Additionally, explicit joint multi-step forecast can be used to estimate the likelihood of rare events---e.g. peak loads, which are critical to safe grid operation---with low variance~\cite{owen2019importance}. 

Success in using predictive models for planning hinges on accurate multi-step trajectories~\cite{nagabandi2018mpc}. 
As described in the previous section, using independent models to forecast different future points in time fails to capture correlations between time points. %, yielding misshaped trajectories. 
Iterative methods are smooth but generally slow to sample from, potentially limiting their effectiveness in sampling-based planners. 

\subsection{Load forecasting}

There is rich and continually expanding literature on probabilistic electric load forecasting~\cite{hong2016probabilistic}. The Global Energy Forecasting competitions have showcased a number of successful techniques for forecasting quantiles of a single-step demand distribution \cite{hong2014global,hong2016gef,hong2019global}. These techniques have included quantile regression, gradient boosted regression trees, quantile random forests, autoregressive integrated moving average, and neural networks. 

More recently, there has been success using long short-term memory (LSTM) \cite{hochreiter1997long}, a form of the recurrent neural network, to do single-step point load forecasting \cite{kong2017short}. 
The LSTM has since been used for single-step probabilistic load forecasting by mapping the hidden state to fixed quantiles of a single-step load distribution and guiding training using the pinball loss with a Huber penalty \cite{wang2019probabilistic}.

Unfortunately, methods that rely on quantile regression cannot be extended to joint multi-step forecasting because there are an infinite number of ways to ascribe $\hat{y}^q_{t+1}$ and $\hat{y}^q_{t+2}$ such that $P(y_{t+1} \leq \hat{y}^q_{t+1} \wedge y_{t+2} \leq \hat{y}^q_{t+2}) = q$. We can however trivially extend neural networks to do joint probabilistic forecasting by using them to parameterize a low-rank, $k$-component Gaussian Mixture Model (GMM) over the multi-step forecast space. That is:
\begin{align}
p(\vect{y}_{t+1:t+K}\mid \vect{y}_{t-L:t}) = \sum_{i=1}^k{\pi_i \mathcal{N}(\vect{y}_{t+1:t+K}; \vect{\mu}_i, \vect{\Sigma}_i)}\text{,}\label{JNN1}\\ 
 \pi_i = \frac{e^{\lambda_i}}{\sum_{j=1}^k{e^{\lambda_j}}}\text{, }
 \vect{\Sigma}_i = \text{diag}(e^{\vect{d}_i})+\vect{B}_i\vect{B}_i^\intercal\text{,}\label{JNN2}\\ 
 (\lambda_i, \vect{\mu}_i, \vect{d}_i, \vect{B}_i)_{i=1}^k = \mathbf{f}(\vect{y}_{t-L:t}; \vect{\theta})\text{,}
\label{JNN3}
\end{align}
where $\pi_i$, $\vect{\mu}_i$, and $\vect{\Sigma}_i$ are the weight, mean, and covariance of each component in the mixture, $B_i \in \mathbb{R}^{K \times r}$ is a matrix with rank $r < K$, and $\mathbf{f}$ is the regression model parameterized by $\vect{\theta}$. This formulation is amenable to use with both feedforward and recurrent neural networks as models $\mathbf{f}$.

The broader field of time-series forecasting is well studied~\cite{hyndman2018forecasting}. 
Methods are typically classified as either using autoregressive (AR)~\cite{billings2013narmax}, or state-space models (SSMs)~\cite{ghahramani1996ldsem,kantas2015particlesurvey,menda2020scalable}. 
While AR models benefit from their versatility and data efficiency, SSMs are more amenable to prior knowledge~\cite{Gupta2020}, and recursive Bayesian estimation.

The Gaussian Process (GP) is a non-parametric approach to describing a distribution over possible functions that map input to output \cite{williams2006gaussian}. 
Gaussian Processes with different kernel functions have been applied to single-step probabilistic electric load forecasting \cite{blum2013electricity,shepero2018residential}. \citet{shepero2018residential} address positive skewness in electricity data by applying a log transform to the input data (rendering a Log-Normal Process). 

The traditional formulation for GPs is single-output, preventing the retrieval of a joint multi-step forecast $p(\vect{y}_{t+1:t+K}\mid\vect{y}_{t-L:t})$. 
Multi-task Gaussian Processes \cite{bonilla2008multi} address this issue by fitting a joint model over all outputs of all data points, using the kernel function
\begin{equation}
\text{Cov}(\vect{y}_{t+i}, \vect{y}'_{t+j}) = 
\mathbf{K}_s(i,j)\mathbf{K}(\vect{y}_{t-L:t},\vect{y'}_{t-L:t})\text{,}
\end{equation}
where $\mathbf{K}_s$ is the scale factor correlating output indices, and $\mathbf{K}$ is a kernel function applied between inputs. 

GPs have also been applied to learn non-parametric state-space models of nonlinear and high-dimensional dynamical systems~\cite{wang2006gpdm,frigola2014variational}. 

%% file: 3-method.tex
\section{Methodologies} \label{sec:method}

This section describes scheduling under uncertainty in a grid environment, describes conditional Gaussian models, and formulates the \textit{Conditional Approximate Normalizing Flow}. 

\subsection{Scheduling under uncertainty}

We consider the problem of choosing the $D$ minimum-load indices from a $K$-step probabilistic forecast in order to schedule some task. 
Given a utility function $U$, we select the ordered set of $D$ time indices $\vect{a}^*_{1:D}$ that maximize utility (and minimize cost):
\begin{equation}
\begin{aligned}
    \vect{a}^*_{1:D} = \text{arg}\max_{\vect{a}_{1:D}}
    \mathbb{M}_{\vect{s}_{1:K}} [U(\vect{s}_{1:K},\vect{a}_{1:D})]\text{,}\\ \vect{s}_{1:K} \sim p(\vect{y}_{t+1:t+K}\mid \vect{y}_{t-L:t})\text{,}\
\end{aligned}
\label{scheduling}
\end{equation}
where $\mathbb{M}$ is an appropriate metric applied onto the utility distribution (e.g., expectation, value-at-risk, conditional value-at-risk~\cite{kochenderfer2019algorithms}). We choose a utility function that sums the forecasted load at scheduled indices, 
\begin{equation}
    U(\vect{s}_{1:K},\vect{a}_{1:D}) = - \sum_{i=1}^D s_{a_i}\text{.}
\label{utility}
\end{equation}
This utility function weights all chosen time indices evenly, but it may be more appropriate to consider an alternative weighting when the allocated indices are not used evenly (e.g. in electric car charging).

\subsection{Conditional Gaussian and Gaussian Mixture Models}\label{cgmm}

Multi-task Gaussian Processes~\cite{bonilla2008multi} can be used to predict correlations between output time indices of single sequences while searching for similarities between different inputs sequences. But they fail to capture the correlation between input and output time indices explicitly. 
One way we can explicitly capture correlations over time between input and output is by fitting an explicit distribution over the full input and target time series, $\vect{y}_{t-L:t+K}$, and conditioning on every newly observed input.
A simple approach is to fit a full-rank multivariate Gaussian to the training data:
\begin{align}
&p(\vect{y}_{t-L:t+K}) = \mathcal{N}(\vect{y}_{t-L:t+K}; \vect{\mu}, \vect{\Sigma})\textit{,}\\
&\vect{\mu} = \frac{1}{n} \sum_{i=1}^n{\vect{y}^i_{t-L:t+K}}\textit{,}\\
&\vect{\Sigma} = \frac{1}{n} \sum_{i=1}^n{(\vect{y}^i_{t-L:t+K}-\vect{\mu})(\vect{y}^{i}_{t-L:t+K}-\vect{\mu})^\intercal}
\textit{,}
\end{align}
with $n$ training data sequences.

Once fit, we can partition our distribution between inputs and outputs and use the Schur complement to form the posterior distribution of a demand trajectory forecast given its input:

\begin{align}
&\begin{bmatrix} \vect{y}_{t-L:t}\\\vect{y}_{t+1:t+K} \end{bmatrix} \sim 
\mathcal{N}\left(
\begin{bmatrix} \vect{\mu}_a\\\vect{\mu}_b \end{bmatrix},
\begin{bmatrix} \vect{\Sigma}_{aa} & \vect{\Sigma}_{ab}\\
                \vect{\Sigma}_{ba} & \vect{\Sigma}_{bb}\\\end{bmatrix}
\right)\text{,}\label{condg1}\\
&p(\vect{y}_{t+1:t+K}\mid \vect{y}_{t-L:t}) = \mathcal{N}(\vect{y}_{t+1:t+K}; \vect{\mu}_{b\mid a}, \vect{\Sigma}_{b\mid a})\text{,}\label{condg2}\\
&\vect{\mu}_{b\mid a} = \vect{\mu}_{b} + \vect{\Sigma}_{ba}\vect{\Sigma}_{aa}^{-1}(\vect{y}_{t-L:t} - \vect{\mu}_{a})\text{,}\label{condg3}\\
&\vect{\Sigma}_{b\mid a} = \vect{\Sigma}_{bb} - \vect{\Sigma}_{ba}\vect{\Sigma}_{aa}^{-1}\vect{\Sigma}_{ab}\text{.}\label{condg4}
\end{align}
This method is a probabilistic analog to linear regression---it can be shown that the predicted mean $\vect{\mu}_{b\mid a}$ is equivalent to that achieved if linear regression were performed to regress to each future index independently.

We can improve upon this method by fitting a richer class of joint distributions to  $p(\vect{y}_{t-L:t+K})$. Again, we choose to fit a $k$-component multivariate Gaussian Mixture Model:
\begin{equation}
p(\vect{y}_{t-L:t+K}) = \sum_{i=1}^k{\pi_i \mathcal{N}(\vect{y}_{t-L:t+K}; \vect{\mu}_i, \vect{\Sigma}_i)}.
\label{GMM1}
\end{equation}
The GMM can be fit easily over training data using the Expectation-Maximization (EM) algorithm, with the number of clusters chosen to minimize the negative log-likelihood of a validation dataset. The GMM is well suited for load trajectory prediction because it can capture multi-modality, which can stem from a number of latent factors (e.g. time-of-day, weekday/weekend/holiday, weather).

Once fit, we can apply a similar method to partition the joint distribution and form an explicit, analytical conditional:
\begin{align}
&\begin{bmatrix} \vect{y}_{t-L:t}\\\vect{y}_{t+1:t+K} \end{bmatrix} \sim 
\sum_{i=1}^k{\pi_i\mathcal{N}\left(
\begin{bmatrix} \vect{\mu}_{ia}\\\vect{\mu}_{ib} \end{bmatrix},
\begin{bmatrix} \vect{\Sigma}_{iaa} & \vect{\Sigma}_{iab}\\
                \vect{\Sigma}_{iba} & \vect{\Sigma}_{ibb}\\\end{bmatrix}
\right)}\text{,}
\label{GMM2}\\
&p(\vect{y}_{t+1:t+K}\mid \vect{y}_{t-L:t}) = \sum_{i=1}^k{\pi'_i \mathcal{N}(\vect{y}_{t+1:t+K}; \vect{\mu}'_i, \vect{\Sigma}'_i)},\label{GMMposterior1}\\
&\pi'_i \propto\pi_i \mathcal{N}(\vect{y}_{t-L:t}; \vect{\mu}_{ia}, \vect{\Sigma}_{iaa}),\label{GMMposterior2}\\
&\vect{\mu}'_i = \vect{\mu}_{ib\mid a} = \vect{\mu}_{ib} + \vect{\Sigma}_{iba}\vect{\Sigma}_{iaa}^{-1}(\vect{y}_{t-L:t} - \vect{\mu}_{ia}),\label{GMMposterior3}\\
&\vect{\Sigma}'_i = \vect{\Sigma}_{ib\mid a} = \vect{\Sigma}_{ibb} - \vect{\Sigma}_{iba}\vect{\Sigma}_{iaa}^{-1}\vect{\Sigma}_{iab}\text{.}
\label{GMMposterior4}
\end{align}
\subsection{Conditional Approximate Normalizing Flow}

One issue with conditional GMMs is that they are limited in the joint distribution $p(\vect{y}_{t-L:t+K})$ they can represent. One alternative for fitting a richer joint distribution is to use a normalizing flow~\cite{rezende2015variational,dinh2016density,kobyzev2020normalizing}. A normalizing flow is a set of reversible transformations that map the input space $\mathcal{Y}$ to some simple distribution (e.g. isotropic Gaussian) on latent space $\mathcal{Z}$. If $f_\theta$ maps $\mathcal{Y}$ to $\mathcal{Z}$, we can express $p_\mathcal{Y}(\vect{y}_{t-L:t+K})$ using the change of variables formula as 
\begin{equation}
    p_\mathcal{Y}(\vect{y}_{t-L:t+K}) = p_\mathcal{Z}(\vect{z}) \left| \det{\frac{\partial f_\theta(\vect{y}_{t-L:t+K})}{\partial \vect{y}_{t-L:t+K}}} \right|\text{.}
\end{equation}
Density estimation (and therefore model fitting) can be done with forward passes through $f_\theta(\vect{y}_{t-L:t+K})$, while sampling can be done efficiently with inverse passes through $f_\theta^{-1}(\vect{z})$.

Unfortunately, we cannot explicitly condition the joint normalizing flow over a partial trajectory. We 
mitigate this by approximating the joint normalizing flow with a many-component GMM, which we can then condition explicitly. The \textit{conditional approximate normalizing flow} (CANF) is formulated as follows:
\begin{enumerate}
    \item Fit a normalizing flow over $p(\vect{y}_{t-L:t+K})$
    \item Approximate the flow by sampling a large number of points and fitting the points with a many-component GMM
    \item To forecast, condition the approximate normalizing flow on partial trajectories with~\cref{GMM1,GMM2,GMMposterior1,GMMposterior2,GMMposterior3,GMMposterior4}
\end{enumerate}
The conditional approximate normalizing flow can alternatively be viewed as a data augmentation method on top of the conditional GMM. In the next section, we see that this method shows strong empirical results.

%% file: 4-experiments.tex
\section{Experiments} \label{sec:experiments}

In our experiments, we demonstrate the efficacy of using an approximate normalizing flow for density estimation, and study the forecast accuracy of our proposed methods on electricity demand data, as well as the quality of decisions to arise from those forecasts.

\subsection{Toy density estimation} \label{sec:toy}

To demonstrate the efficacy of using an approximate normalizing flow to fit a joint distribution over $\vect{y}_{t-L:t+K}$, we compare the use of a Gaussian mixture model, normalizing flow, and approximate normalizing flow to fit samples from $p_\text{data}=\mathcal{U}([0,1)^2)$. We fit each model $p_\theta$ using 1000 samples for training by maximum likelihood estimation and 200 samples for validation to select hyperparameters. Our validation process find that the best GMM has 9 components, the best RealNVP~\cite{dinh2016density} normalizing flow model uses 4 coupling layers which each use 2-layer, 12-unit feedforward networks for scale and translation, and the best final model approximates the normalizing flow by sampling 10k points and fitting a 40-component GMM.

\begin{figure*}[h]
  \centering
  \subcaptionbox{GMM}[.3\linewidth][c]{%
    \includegraphics[trim=52 32 52 32,clip,width=\linewidth]{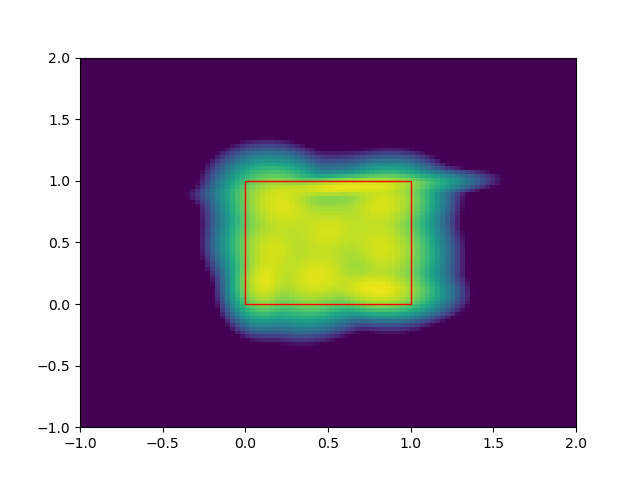}}\quad
  \subcaptionbox{Normalizing flow}[.3\linewidth][c]{%
    \includegraphics[trim=52 32 52 32,clip,width=\linewidth]{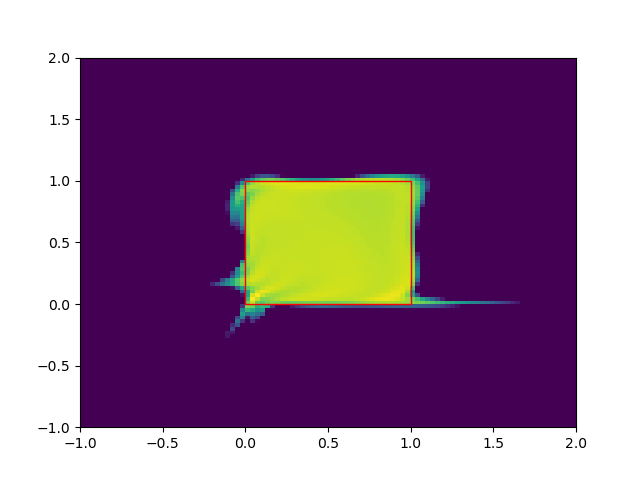}}\quad
  \subcaptionbox{Approx. normalizing flow}[.3\linewidth][c]{%
    \includegraphics[trim=52 32 52 32,clip,width=\linewidth]{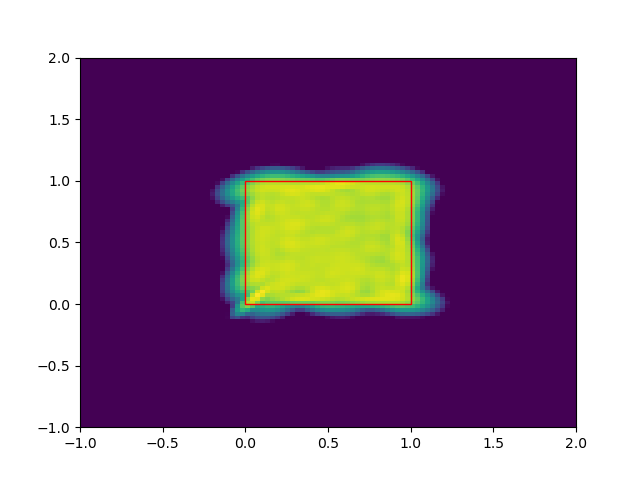}}
\caption{Log-density estimates when using GMM, normalizing flow, and approximate normalizing flow models to fit 1000 samples from $\mathcal{U}([0,1)^2)$ (in red).}
    \label{fig:toy}
\end{figure*}

In~\Cref{fig:toy}, we can see the fitted distributions compared to the underlying sample-generating box. We see that the normalizing flow model fits the joint density better than the GMM. When we approximate the normalizing flow (in order to be able to condition explicitly), though we lose estimation quality compared to the normalizing flow, we still see a net improvement compared to the GMM. We compare Kullback-Leibler divergences $D_{KL}(p_\text{data} \parallel p_\theta)=\mathbb{E}_{\vect{x} \sim p_\text{data}(\cdot)} [\log p_\text{data}(\vect{x}) - \log p_\theta(\vect{x})]$ after training each model ten times with the tuned hyperparameters. We find the GMM, normalizing flow, and approximate normalizing flow models exhibit mean divergences of $0.150 \pm 0.002$, $0.082 \pm 0.004$, and $0.102 \pm 0.005$ respectively. \textit{That is, the approximate normalizing flow yields a one-third improvement over the original GMM for estimating the joint density function while remaining amenable to explicit conditioning.}

\subsection{Demand forecasting}

We evaluate our proposed methods for joint multi-step demand forecasting on the `OpenEI Commercial and Residential Hourly Load Profiles for all TMY3 Locations in the United States' dataset~\cite{OPENEI}, and compare performance against iterative single-step forecast rollouts. We focus on forecasting demand over the next 12 hours ($K=12$) given either demand from the past 8 hours ($L=7$) or the past 24 hours ($L=23$). We believe that forecasting 12 hours is appropriate for a range of scheduling applications, and that 8 and 24 hours are reasonable input lengths. We compare methods at two arbitrarily chosen locations: Bakersfield, California, and Salt Lake City, Utah.

\subsubsection{Dataset} 
The OpenEI Commercial and Residential Hourly Load Profiles dataset includes a year of hourly load consumption data for residential houses in various locations within the US. The houses in the dataset are tailored to represent what might be expected from a typical house in that location. Though the consumption patterns in the dataset are on the household level, their smoothness is representative of what one might also expect at a higher level in the energy hierarchy. 
Since we only have one year of data and seasonality plays a factor in consumption, we cannot choose our testing data as the latter part of the time series.
Instead, we randomly select 25$\%$ of weeks to form our testing set. Then, we apply a rolling window to our training and our testing sets separately to compile every $(K+L+1)$-step input-output time sequence in each set.

Consumption forecasting methods can be improved by taking into account weather or time (e.g. hour, weekday, month) information. We stray from doing so here in order to focus our comparison on assessing the relative difference between methods given past demand alone. In practice, accounting for additional information sources would be straightforward. For almost all methods listed here, we can simply append the information embedding vector $\vect{e}_t$ to the input sequence vector $\vect{y}_{t-L:t}$ at train and test time.\footnote{For recurrent neural networks, the embedding should be appended to the warmed-up hidden state $\vect{h}_t$.}

\begin{table*}[ht]
\ra{1.2}
\caption{Multi-step forecast performance and decision scores with both via $K$-step (I)terative rollouts and explicit (J)oint forecast models for $p(\vect{y}_{t+1:t+12}\mid\vect{y}_{t-7:t})$ (above) and $p(\vect{y}_{t+1:t+12}\mid\vect{y}_{t-23:t})$ (below). Means and standard deviations are computed from up to 10 successful trials. }
\centering
\scalebox{0.72}{\input{figs/datatable}}
\label{table:results}
\end{table*}

\subsubsection{Evaluation metrics} 
We use weighted absolute percentage error (WAPE) and root-weighted-square error (RWSE) to evaluate variability in deviations from the true demand trajectories under the predictive probability distributions:
\begin{align}
    &WAPE = &\frac{1}{nmK}\sum_{i=1}^n{\sum_{\tau=1}^K{\sum_{j=1}^m{\mid\frac{y^i_{t+\tau} - z^{ij}_{t+\tau}}{y^i_{t+\tau}}\mid}}},\\
    &RWSE = &\sqrt{\frac{1}{nmK}\sum_{i=1}^n{\sum_{\tau=1}^K{\sum_{j=1}^m{\big(y^i_{t+\tau} - z^{ij}_{t+\tau}\big)^2}}}},\\
    &&\vect{z}^{ij}_{t+1:t+K} \sim p(\vect{y}^i_{t+1:t+K}\mid\vect{y}^i_{t-L:t}) \nonumber \\
    &&\forall\  i \in \{1,2,...,n\}, j \in \{1,2,...,m\}\text{,}\nonumber
\end{align}
where $n$ is the number of testing sequences, $m$ is a large number of samples (1000) from the predictive distribution, and $\vect{z}^{ij}_{t+1:t+K}$ is the $j$th demand trajectory sample from the predictive distribution of the $i$th test sequence. These metrics quantify linear- and quadratic-like errors for probabilistic forecasts. 
Unfortunately, metrics that score single-step probabilistic forecasts based on their quantiles cannot be extended to higher dimensions (e.g., pinball loss). 
Where applicable, we also compare the mean log-likelihood of the true trajectories under the predicted distribution during testing (LL).

To compare decision quality, we use our predictive forecasts to predict $D=4$ indices in which load is minimized. 
We choose the action that maximizes the 20\% value-at-risk on utility (minimizing the 80\% value-at-risk on cost) using~\Cref{scheduling,utility}.

We report a decision score which is the 80\% quantile on \textit{proportional regret} across the test set. Proportional regret is defined as
\begin{equation*}
    PR = \frac{U(\vect{y}_{t+1:t+K}, \vect{a}_{1:D}) - U(\vect{y}_{t+1:t+K}, \vect{a}^*_{1:D})}{U(\vect{y}_{t+1:t+K}, \vect{a}^*_{1:D})}\text{,}
\end{equation*}
where $\vect{y}_{t+1:t+K}$ and $\vect{a}^*_{1:D}$ are the true future load and best action in hindsight and $\vect{a}_{1:D}$ is the selected action.
The decision score therefore indicates the bound on suboptimality achieved 80\% of the time.

\subsubsection{Iterative forecast methods} To compare joint demand forecasting with iterative rollouts of single-step forecasts, we fit a number of single-step models. We obtain samples from the $K$-step joint distribution by iteratively sampling from the models and updating their inputs.

\textit{ARMA:} Autoregressive moving average (ARMA) predicts the demand at the next step to be normally distributed around a linear combination of the previous steps. The single-step ARMA model can be found by applying the conditional Gaussian method to single-step forecasting.

\textit{IFNN/IRNN:} The single-step feedforward neural network (IFNN) uses a 3-layer, 40-unit/layer neural network to map to a 3-component univariate Gaussian Mixture Model. The single-step recurrent neural network (IRNN) uses an LSTM block with 40 hidden units and a 3-layer output network with 20 units/layer and map to a 3-component univariate Gaussian Mixture Model. This is close to what was done by \citet{wang2019probabilistic}, who mapped an LSTM to single-step quantiles trained to minimize pinball loss rather than to a GMM trained to minimize NLL. 

\subsubsection{Joint forecast methods}Implementation of the conditional Gaussian model (CG) is straightforward, and described in \Cref{condg1,condg2,condg3,condg4}. To fit a GMM over the joint input-output distribution to use in the conditional GMM model (CGMM), we use the Scikit-learn implementation of EM \cite{scikit} and five components in the mixture. 
To fit the multi-output Gaussian Process (MOGP) model, we use the Gpytorch implementation of multi-task GPs \cite{gardner2018gpytorch} with a rank-5 inter-output scale kernel $\mathbf{K}_s$. Our input kernel $\mathbf{K}$ is a product of a Mat\'{e}rn kernel and a rational quadratic kernel, both with index-independent characteristic length scales.

For the joint feedforward neural network model (JFNN, \Cref{JNN1,JNN2,JNN3}), we use a 3 layer, 40 unit/layer network to map to a two-component GMM, where all covariances are rank-2. For the joint RNN model (JRNN, \Cref{JNN1,JNN2,JNN3}), we use an LSTM to map inputs to a 40-unit hidden state. The hidden state is mapped to a 3-component GMM with rank-3 covariance matrices through a 3 layer, 40 unit/layer feedforward neural network. 

For the conditional approximate normalizing flow model (CANF), we use RealNVP~\cite{dinh2016density} with 10 coupling layers which each use 2-layer, 32-unit feedforward networks for scale and translation within the coupling. Once trained, we sample one million points, and fit a 25-component GMM to approximate the flow.

All parameterized models are trained using the Adam optimizer~\cite{kingma2014adam} to minimize the negative log-likelihood of future loads given their predictive distributions. 

\subsection{Results}

In order to compare multi-step forecast accuracy and decision quality between all the methods, we report metrics on forecasts of $\vect{y}_{t+1:t+12}\mid\vect{y}_{t-7:t}$ and $\vect{y}_{t+1:t+12}\mid\vect{y}_{t-23:t}$ in \cref{table:results}. We see that the joint forecasting methods are indeed of similar or even of better quality than iterative ones, although they require predicting more distributional parameters. As expected, we can see a correlation between forecast accuracy (in terms of WAPE, RWSE, and LL) and decision score.

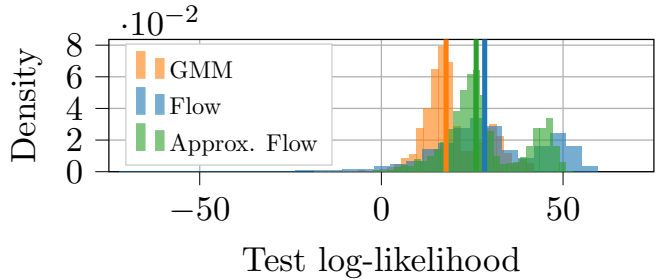
\begin{figure}[tbp]
\centerline{\scalebox{1.20}{\input{figs/likelihoods.tex}}}
\caption{Distributions over full trajectory log-likelihood, $\log p(\vect{y}_{t-7:t+12})$, achieved under GMM, normalizing flow, and approximated normalizing flow models on the test set in Bakersfield, CA.}
\label{fig:likelihood}
\end{figure}

As shown in \cref{table:results}, CANF can achieve superior performance in terms of both forecast accuracy and most importantly, decision quality. In the Bakersfield $\vect{y}_{t+1:t+12}\mid\vect{y}_{t-7:t}$ setting for example, the CANF yields a RWSE that is 34\% better than the CGMM (the next best joint method), 41\% better than the IFNN (the next best iterative method), and decision scores that are ten time better than both alternatives. Furthermore, when looking at the distribution over log-likelihoods of full test sequences (\cref{fig:likelihood}), we observe approximate flows are much better at GMMs at characterizing the full trajectory likelihoods and are only slightly worse than their non-approximate counterparts. This result matches that of~\Cref{sec:toy}. 
We also find that non-approximate normalizing flows are prone to having adverse outliers from the flow models overfitting, but this behavior gets curtailed when approximating.

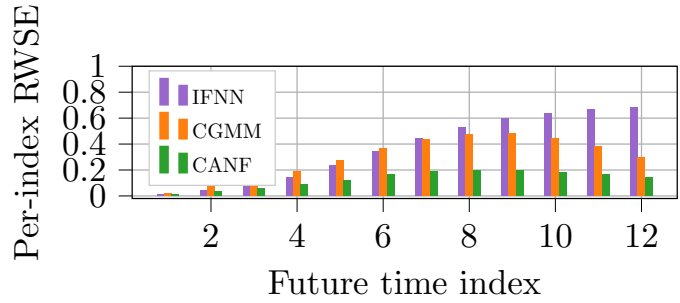
\begin{figure}[tbp]
\centerline{\scalebox{1.20}{\input{figs/errors.tex}}}
%\centerline{\scalebox{0.6}{\includegraphics{figs/errors.png}}}
\caption{Evolution of per-index RWSE when using the IFNN, CGMM, and CANF models to predict $\vect{y}_{t+1:t+12}\mid\vect{y}_{t-7:t}$ in Bakersfield, CA.}
\label{fig:errors}
\end{figure}

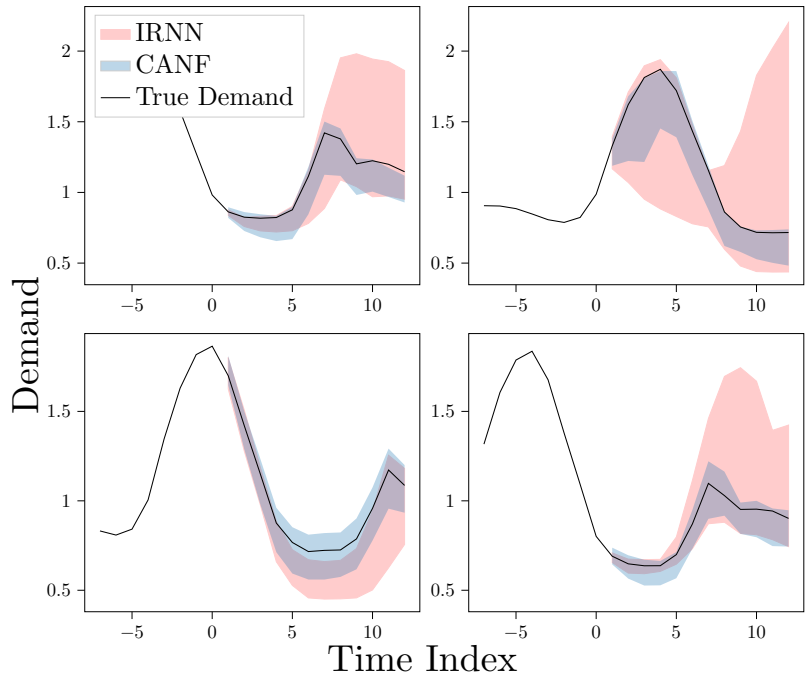
\begin{figure}[tbp]
\centerline{\scalebox{0.65}{\input{figs/samples.tex}}}
%\centerline{\scalebox{0.6}{\includegraphics{figs/samples.png}}}
% note: original size 3.8in
\caption{Sample multi-step forecast distributions predicting $\vect{y}_{t+1:t+12}\mid\vect{y}_{t-7:t}$ in Salt Lake City, UT using the IRNN and CANF models.}
\label{fig:samples}
\end{figure}

We examine this performance further by measuring the weighted error metrics per time-index to determine how the quality of sampled trajectories evolves. The per-index RWSE using the IFNN, CGMM, and CANF models to forecast $\vect{y}_{t+1:t+12} \mid \vect{y}_{t-7:t}$ is presented in \cref{fig:errors}.

From this per-index break-down, it becomes clear that the joint methods capture longer time-horizon correlations that the rollouts do not (in this case, the cyclical nature of electricity demand). This is because, typically, iterative methods cannot capture long-horizon correlations and accumulate errors over time, where this is not an issue for joint methods.

We show some samples from rollouts of the single-step RNN overlaid with the demand distribution explicitly predicted using a conditional approximate normalizing flow at random test points in \cref{fig:samples}. As expected, the forecast uncertainty remains bounded for CANF but grows with time for IRNN.

The experiments conducted empirically support the claims that conditional approximate normalizing flows give superior performance in terms of forecast accuracy and decision quality for scheduling in the explored setting.

%% file: figs/datatable.tex
\begin{tabular}{@{}rr RRRR r RRRR @{}}
\toprule
&&\multicolumn{4}{c}{\textbf{Bakersfield, CA}} & \phantom{a}
&\multicolumn{4}{c}{\textbf{Salt Lake City, UT}}\\
\cmidrule{3-6} \cmidrule{8-11}
&\textbf{Model} &  
{\textbf{WAPE}}  & {\textbf{RWSE}} & {\textbf{LL}} & {\textbf{Dec. Score}} &&
{\textbf{WAPE}}  & {\textbf{RWSE}} & {\textbf{LL}} & {\textbf{Dec. Score}} \\
\midrule
\multirow{9}{*}{\rotatebox{90}{$p(\vect{y}_{t+1:t+12}\mid\vect{y}_{t-7:t})$}}
&\texttt{ARMA} (I)            & 0.529 \phantom{\scriptstyle\pm 0.000 }& 0.746\phantom{\scriptstyle\pm 0.000 } & \text{---} & 0.384 \phantom{\scriptstyle\pm 0.000 }&
                    & 0.460 \phantom{\scriptstyle\pm 0.000 }& 0.537 \phantom{\scriptstyle\pm 0.000 }& \text{---} & 0.396\phantom{\scriptstyle\pm 0.000 }\\
&\texttt{IFNN} (I)            & 0.310 {\scriptstyle\pm 0.068 }& 0.596 {\scriptstyle\pm 0.047 }& \text{---} & 0.121 {\scriptstyle\pm 0.078 }&
                    & 0.230 {\scriptstyle\pm 0.026 }& 0.432 {\scriptstyle\pm 0.146 }& \text{---} & 0.106 {\scriptstyle\pm 0.045}\\
&\texttt{IRNN} (I)            & 0.311 {\scriptstyle\pm 0.051 }& 0.620 {\scriptstyle\pm 0.014 }& \text{---} & 0.185 {\scriptstyle\pm 0.047 }&
                    & 0.262 {\scriptstyle\pm 0.036 }& 0.399 {\scriptstyle\pm 0.036 }& \text{---} & 0.163 {\scriptstyle\pm 0.073}\\
&\texttt{CG} (J)           & 0.521\phantom{\scriptstyle\pm 0.000 } & 0.716\phantom{\scriptstyle\pm 0.000 } & 10.4\phantom{\scriptstyle\pm 0.0 } & 0.340 \phantom{\scriptstyle\pm 0.000 }&
                    & 0.431 \phantom{\scriptstyle\pm 0.000 }& 0.485 \phantom{\scriptstyle\pm 0.000 }& 11.2 \phantom{\scriptstyle\pm 0.0 }& 0.323\phantom{\scriptstyle\pm 0.000 }\\
&\texttt{JFNN} (J)            & 0.358 {\scriptstyle\pm 0.025 }& 0.651 {\scriptstyle\pm 0.030 }& 14.6 {\scriptstyle\pm 0.6 }& 0.101 {\scriptstyle\pm 0.023 }&
                    & 0.262 {\scriptstyle\pm 0.018 }& 0.398 {\scriptstyle\pm 0.017 }& 12.4 {\scriptstyle\pm 1.7 }& 0.085 {\scriptstyle\pm 0.022}\\
&\texttt{JRNN} (J)            & 0.269 {\scriptstyle\pm 0.023 }& 0.543 {\scriptstyle\pm 0.048 }& 14.5 {\scriptstyle\pm 1.1 }& 0.030 {\scriptstyle\pm 0.011 }&
                    & 0.214 {\scriptstyle\pm 0.015 }& 0.340 {\scriptstyle\pm 0.017 }& 13.9 {\scriptstyle\pm 1.6 }& 0.046 {\scriptstyle\pm 0.014}\\
&\texttt{MOGP} (J)            & 0.559 {\scriptstyle\pm 0.043 }& 0.770 {\scriptstyle\pm 0.050 }& -1.3 {\scriptstyle\pm 0.6 }& 0.144 {\scriptstyle\pm 0.051 }&
                    & 0.431 {\scriptstyle\pm 0.016 }& 0.504 {\scriptstyle\pm 0.017 }& 0.6 {\scriptstyle\pm 0.5 }& 0.120 {\scriptstyle\pm 0.013}\\
&\texttt{CGMM} (J)         & 0.321 {\scriptstyle\pm 0.013 }& 0.536 {\scriptstyle\pm 0.024 }& 15.4 {\scriptstyle\pm 0.1 }& 0.131 {\scriptstyle\pm 0.010 }&
                    & 0.217 {\scriptstyle\pm 0.003 }& 0.344 {\scriptstyle\pm 0.009 }& 14.4 {\scriptstyle\pm 0.2 }& 0.077 {\scriptstyle\pm 0.004}\\
&\texttt{CANF} (J)            & \mathbf{0.192} {\scriptstyle\pm 0.007 }& \mathbf{0.352} {\scriptstyle\pm 0.013 }& \mathbf{20.6} {\scriptstyle\pm 0.3 }& \mathbf{0.012} {\scriptstyle\pm 0.001 }&
                    & \mathbf{0.173} {\scriptstyle\pm 0.010 }& \mathbf{0.295} {\scriptstyle\pm 0.011 }& \mathbf{21.9} {\scriptstyle\pm 0.5 }& \mathbf{0.037} {\scriptstyle\pm 0.013}\\
\midrule
\multirow{9}{*}{\rotatebox{90}{$p(\vect{y}_{t+1:t+12}\mid\vect{y}_{t-23:t})$}}
&\texttt{ARMA} (I)            & 0.163 \phantom{\scriptstyle\pm 0.000 }& 0.218 \phantom{\scriptstyle\pm 0.000 }& \text{---} & 0.0078 \phantom{\scriptstyle\pm 0.0000 }&
                    & 0.109 \phantom{\scriptstyle\pm 0.000 }& 0.149 \phantom{\scriptstyle\pm 0.000 }& \text{---} & 0.0079\phantom{\scriptstyle\pm 0.0000 }\\
&\texttt{IFNN} (I)            & 0.123 {\scriptstyle\pm 0.008 }& 0.261 {\scriptstyle\pm 0.025 }& \text{---} & 0.0041 {\scriptstyle\pm 0.0015 }&
                    & 0.103 {\scriptstyle\pm 0.025 }& 0.169 {\scriptstyle\pm 0.021 }& \text{---} & 0.0075 {\scriptstyle\pm 0.0061}\\
&\texttt{IRNN} (I)            & 0.216 {\scriptstyle\pm 0.069 }& 0.371 {\scriptstyle\pm 0.094 }& \text{---} & 0.0605 {\scriptstyle\pm 0.0359 }&
                    & 0.171 {\scriptstyle\pm 0.033 }& 0.258 {\scriptstyle\pm 0.043 }& \text{---} & 0.0400 {\scriptstyle\pm 0.0265}\\
&\texttt{CG} (J)           & 0.150 \phantom{\scriptstyle\pm 0.000 }& 0.199 \phantom{\scriptstyle\pm 0.000 }& 19.0 \phantom{\scriptstyle\pm 0.0 }& 0.0057 \phantom{\scriptstyle\pm 0.0000 }&
                    & 0.103 \phantom{\scriptstyle\pm 0.000 }& 0.139 \phantom{\scriptstyle\pm 0.000 }& 16.6 \phantom{\scriptstyle\pm 0.0 }& 0.0037\phantom{\scriptstyle\pm 0.0000 }\\
&\texttt{JFNN} (J)            & 0.144 {\scriptstyle\pm 0.006 }& 0.269 {\scriptstyle\pm 0.006 }& 19.5 {\scriptstyle\pm 0.3 }& 0.0043 {\scriptstyle\pm 0.0012 }&
                    & 0.135 {\scriptstyle\pm 0.039 }& 0.205 {\scriptstyle\pm 0.041 }& 19.6 {\scriptstyle\pm 1.2 }& 0.0107 {\scriptstyle\pm 0.0110}\\
&\texttt{JRNN} (J)            & 0.191 {\scriptstyle\pm 0.027 }& 0.394 {\scriptstyle\pm 0.070 }& 17.7 {\scriptstyle\pm 0.6 }& 0.0092 {\scriptstyle\pm 0.0074 }&
                    & 0.135 {\scriptstyle\pm 0.009 }& 0.219 {\scriptstyle\pm 0.013 }& 18.3 {\scriptstyle\pm 0.8 }& 0.0096 {\scriptstyle\pm 0.0009}\\
&\texttt{MOGP} (J)            & 0.197 {\scriptstyle\pm 0.029 }& 0.282 {\scriptstyle\pm 0.022 }& 7.0 {\scriptstyle\pm 0.9 }& 0.0181 {\scriptstyle\pm 0.0146 }&
                    & 0.150 {\scriptstyle\pm 0.014 }& 0.183 {\scriptstyle\pm 0.011 }& 11.1 {\scriptstyle\pm 1.5 }& 0.0170 {\scriptstyle\pm 0.0159}\\
&\texttt{CGMM} (J)         & 0.125 {\scriptstyle\pm 0.006 }& 0.196 {\scriptstyle\pm 0.008 }& 22.2 {\scriptstyle\pm 0.8 }& 0.0041 {\scriptstyle\pm 0.0011 }&
                    & \mathbf{0.087} {\scriptstyle\pm 0.000 }& \mathbf{0.130} {\scriptstyle\pm 0.001 }& 18.6 {\scriptstyle\pm 0.1 }& 0.0068 {\scriptstyle\pm 0.0006}\\
&\texttt{CANF} (J)            & \mathbf{0.107} {\scriptstyle\pm 0.001 }& \mathbf{0.186} {\scriptstyle\pm 0.004 }& \mathbf{23.2} {\scriptstyle\pm 0.4 }& \mathbf{0.0013} {\scriptstyle\pm 0.0009 }&
                    & \mathbf{0.088} {\scriptstyle\pm 0.003 }& 0.137 {\scriptstyle\pm 0.003 }& \mathbf{25.4} {\scriptstyle\pm 0.4 }& \mathbf{0.0034} {\scriptstyle\pm 0.0010}\\
\bottomrule
\end{tabular}

%% file: figs/likelihoods.tex
% This file was created by matplotlib2tikz v0.7.5.
\begin{tikzpicture}

\definecolor{color1}{rgb}{0.12156862745098,0.466666666666667,0.705882352941177}
\definecolor{color0}{rgb}{1,0.498039215686275,0.0549019607843137}
\definecolor{color2}{rgb}{0.172549019607843,0.627450980392157,0.172549019607843}

\begin{axis}[
legend cell align={left},
legend style={at={(0.03,0.97)},
nodes={scale=0.7, transform shape},
anchor=north west, draw=white!80.0!black},
tick align=outside,
tick pos=left,
x grid style={white!69.01960784313725!black},
xlabel={Test log-likelihood},
xmajorgrids,
xmin=-75, xmax=75,
xtick style={color=black},
y grid style={white!69.01960784313725!black},
ylabel={Density},
ymajorgrids,
ymin=0, ymax=0.0836076969644542,
ytick style={color=black},
width=0.5\textwidth,
height=0.2\textwidth
]
\draw[fill=color0,draw opacity=0,fill opacity=0.6] (axis cs:-18.5130875127577,0) rectangle (axis cs:-16.5021689208631,0.000216965607796689);
\addlegendimage{ybar,ybar legend,fill=color0,draw opacity=0,fill opacity=0.6};
\addlegendentry{GMM}

\draw[fill=color0,draw opacity=0,fill opacity=0.6] (axis cs:-16.5021689208631,0) rectangle (axis cs:-14.4912503289685,0);
\draw[fill=color0,draw opacity=0,fill opacity=0.6] (axis cs:-14.4912503289685,0) rectangle (axis cs:-12.480331737074,0.000433931215593378);
\draw[fill=color0,draw opacity=0,fill opacity=0.6] (axis cs:-12.4803317370739,0) rectangle (axis cs:-10.4694131451794,0.000216965607796689);
\draw[fill=color0,draw opacity=0,fill opacity=0.6] (axis cs:-10.4694131451794,0) rectangle (axis cs:-8.45849455328479,0.000216965607796689);
\draw[fill=color0,draw opacity=0,fill opacity=0.6] (axis cs:-8.45849455328479,0) rectangle (axis cs:-6.44757596139021,0.000433931215593378);
\draw[fill=color0,draw opacity=0,fill opacity=0.6] (axis cs:-6.44757596139021,0) rectangle (axis cs:-4.43665736949563,0.000650896823390067);
\draw[fill=color0,draw opacity=0,fill opacity=0.6] (axis cs:-4.43665736949563,0) rectangle (axis cs:-2.42573877760104,0.000216965607796689);
\draw[fill=color0,draw opacity=0,fill opacity=0.6] (axis cs:-2.42573877760104,0) rectangle (axis cs:-0.414820185706464,0.000867862431186757);
\draw[fill=color0,draw opacity=0,fill opacity=0.6] (axis cs:-0.414820185706464,0) rectangle (axis cs:1.59609840618812,0.00173572486237351);
\draw[fill=color0,draw opacity=0,fill opacity=0.6] (axis cs:1.59609840618812,0) rectangle (axis cs:3.6070169980827,0.00282055290135695);
\draw[fill=color0,draw opacity=0,fill opacity=0.6] (axis cs:3.6070169980827,0) rectangle (axis cs:5.61793558997728,0.00455627776373047);
\draw[fill=color0,draw opacity=0,fill opacity=0.6] (axis cs:5.61793558997728,0) rectangle (axis cs:7.62885418187186,0.00846165870407088);
\draw[fill=color0,draw opacity=0,fill opacity=0.6] (axis cs:7.62885418187186,0) rectangle (axis cs:9.63977277376645,0.0110652459976311);
\draw[fill=color0,draw opacity=0,fill opacity=0.6] (axis cs:9.63977277376645,0) rectangle (axis cs:11.650691365661,0.0277715977979762);
\draw[fill=color0,draw opacity=0,fill opacity=0.6] (axis cs:11.650691365661,0) rectangle (axis cs:13.6616099575556,0.0399216718345908);
\draw[fill=color0,draw opacity=0,fill opacity=0.6] (axis cs:13.6616099575556,0) rectangle (axis cs:15.6725285494502,0.0646557511234134);
\draw[fill=color0,draw opacity=0,fill opacity=0.6] (axis cs:15.6725285494502,0) rectangle (axis cs:17.6834471413448,0.0796263780613849);
\draw[fill=color0,draw opacity=0,fill opacity=0.6] (axis cs:17.6834471413448,0) rectangle (axis cs:19.6943657332394,0.0689950632793469);
\draw[fill=color0,draw opacity=0,fill opacity=0.6] (axis cs:19.6943657332394,0) rectangle (axis cs:21.7052843251339,0.0288564258369597);
\draw[fill=color0,draw opacity=0,fill opacity=0.6] (axis cs:21.7052843251339,0) rectangle (axis cs:23.7162029170285,0.0130179364678013);
\draw[fill=color0,draw opacity=0,fill opacity=0.6] (axis cs:23.7162029170285,0) rectangle (axis cs:25.7271215089231,0.0128009708600046);
\draw[fill=color0,draw opacity=0,fill opacity=0.6] (axis cs:25.7271215089231,0) rectangle (axis cs:27.7380401008177,0.0186590422705153);
\draw[fill=color0,draw opacity=0,fill opacity=0.6] (axis cs:27.7380401008177,0) rectangle (axis cs:29.7489586927123,0.0277715977979762);
\draw[fill=color0,draw opacity=0,fill opacity=0.6] (axis cs:29.7489586927123,0) rectangle (axis cs:31.7598772846068,0.0301582194837398);
\draw[fill=color0,draw opacity=0,fill opacity=0.6] (axis cs:31.7598772846068,0) rectangle (axis cs:33.7707958765014,0.0225644232108556);
\draw[fill=color0,draw opacity=0,fill opacity=0.6] (axis cs:33.7707958765014,0) rectangle (axis cs:35.781714468396,0.0119331084288179);
\draw[fill=color0,draw opacity=0,fill opacity=0.6] (axis cs:35.781714468396,0) rectangle (axis cs:37.7926330602906,0.00520717458712054);
\draw[fill=color0,draw opacity=0,fill opacity=0.6] (axis cs:37.7926330602906,0) rectangle (axis cs:39.8035516521852,0.00781076188068081);
\draw[fill=color0,draw opacity=0,fill opacity=0.6] (axis cs:39.8035516521852,0) rectangle (axis cs:41.8144702440798,0.00564110580271388);
\draw[fill=color1,draw opacity=0,fill opacity=0.6] (axis cs:-72.1333770751953,0) rectangle (axis cs:-67.7454605102539,0.00019886438954014);
\addlegendimage{ybar,ybar legend,fill=color1,draw opacity=0,fill opacity=0.6};
\addlegendentry{Flow}

\draw[fill=color1,draw opacity=0,fill opacity=0.6] (axis cs:-67.7454681396484,0) rectangle (axis cs:-63.3575592041016,0.000198864562425857);
\draw[fill=color1,draw opacity=0,fill opacity=0.6] (axis cs:-63.357551574707,0) rectangle (axis cs:-58.9696426391602,9.94322812129285e-05);
\draw[fill=color1,draw opacity=0,fill opacity=0.6] (axis cs:-58.9696426391602,0) rectangle (axis cs:-54.5817260742188,9.94321947700699e-05);
\draw[fill=color1,draw opacity=0,fill opacity=0.6] (axis cs:-54.5817260742188,0) rectangle (axis cs:-50.1938171386719,0);
\draw[fill=color1,draw opacity=0,fill opacity=0.6] (axis cs:-50.1938095092773,0) rectangle (axis cs:-45.8059005737305,0.000497161406064642);
\draw[fill=color1,draw opacity=0,fill opacity=0.6] (axis cs:-45.8059005737305,0) rectangle (axis cs:-41.4179840087891,0.00019886438954014);
\draw[fill=color1,draw opacity=0,fill opacity=0.6] (axis cs:-41.4179840087891,0) rectangle (axis cs:-37.0300750732422,0.000198864562425857);
\draw[fill=color1,draw opacity=0,fill opacity=0.6] (axis cs:-37.0300712585449,0) rectangle (axis cs:-32.6421546936035,0.00019886438954014);
\draw[fill=color1,draw opacity=0,fill opacity=0.6] (axis cs:-32.6421508789062,0) rectangle (axis cs:-28.2542419433594,0.000596593687277571);
\draw[fill=color1,draw opacity=0,fill opacity=0.6] (axis cs:-28.2542419433594,0) rectangle (axis cs:-23.866325378418,0.000198864475982961);
\draw[fill=color1,draw opacity=0,fill opacity=0.6] (axis cs:-23.8663291931152,0) rectangle (axis cs:-19.4784126281738,0.000894890141923324);
\draw[fill=color1,draw opacity=0,fill opacity=0.6] (axis cs:-19.4784145355225,0) rectangle (axis cs:-15.0904979705811,0.000795458076817598);
\draw[fill=color1,draw opacity=0,fill opacity=0.6] (axis cs:-15.0904998779297,0) rectangle (axis cs:-10.7025833129883,0.000696025817215398);
\draw[fill=color1,draw opacity=0,fill opacity=0.6] (axis cs:-10.7025861740112,0) rectangle (axis cs:-6.31466960906982,0.00129261909388924);
\draw[fill=color1,draw opacity=0,fill opacity=0.6] (axis cs:-6.31467151641846,0) rectangle (axis cs:-1.92675495147705,0.00188921272713994);
\draw[fill=color1,draw opacity=0,fill opacity=0.6] (axis cs:-1.92675745487213,0) rectangle (axis cs:2.46115922927856,0.00338069645909252);
\draw[fill=color1,draw opacity=0,fill opacity=0.6] (axis cs:2.46115684509277,0) rectangle (axis cs:6.84907341003418,0.00467331518559958);
\draw[fill=color1,draw opacity=0,fill opacity=0.6] (axis cs:6.84907150268555,0) rectangle (axis cs:11.236988067627,0.00646309687414298);
\draw[fill=color1,draw opacity=0,fill opacity=0.6] (axis cs:11.236985206604,0) rectangle (axis cs:15.6249017715454,0.0143182422707732);
\draw[fill=color1,draw opacity=0,fill opacity=0.6] (axis cs:15.6249008178711,0) rectangle (axis cs:20.0128173828125,0.0180966673144494);
\draw[fill=color1,draw opacity=0,fill opacity=0.6] (axis cs:20.0128135681152,0) rectangle (axis cs:24.4007301330566,0.0274432976856486);
\draw[fill=color1,draw opacity=0,fill opacity=0.6] (axis cs:24.4007320404053,0) rectangle (axis cs:28.7886409759521,0.0340058401748215);
\draw[fill=color1,draw opacity=0,fill opacity=0.6] (axis cs:28.7886428833008,0) rectangle (axis cs:33.1765594482422,0.0287359042885502);
\draw[fill=color1,draw opacity=0,fill opacity=0.6] (axis cs:33.1765594482422,0) rectangle (axis cs:37.5644683837891,0.0138210870885971);
\draw[fill=color1,draw opacity=0,fill opacity=0.6] (axis cs:37.5644760131836,0) rectangle (axis cs:41.9523849487305,0.00825287934067306);
\draw[fill=color1,draw opacity=0,fill opacity=0.6] (axis cs:41.9523849487305,0) rectangle (axis cs:46.3403015136719,0.0161080155527513);
\draw[fill=color1,draw opacity=0,fill opacity=0.6] (axis cs:46.3403015136719,0) rectangle (axis cs:50.7282104492188,0.0244603411783804);
\draw[fill=color1,draw opacity=0,fill opacity=0.6] (axis cs:50.728214263916,0) rectangle (axis cs:55.1161308288574,0.0163068799422915);
\draw[fill=color1,draw opacity=0,fill opacity=0.6] (axis cs:55.1161346435547,0) rectangle (axis cs:59.5040435791016,0.00377842668609128);
\draw[fill=color2,draw opacity=0,fill opacity=0.6] (axis cs:-1.95284055273715,0) rectangle (axis cs:-0.132023309975551,0.000718853321915508);
\addlegendimage{ybar,ybar legend,fill=color2,draw opacity=0,fill opacity=0.6};
\addlegendentry{Approx. Flow}

\draw[fill=color2,draw opacity=0,fill opacity=0.6] (axis cs:-0.132023309975551,0) rectangle (axis cs:1.68879393278605,0.000239617773971836);
\draw[fill=color2,draw opacity=0,fill opacity=0.6] (axis cs:1.68879393278605,0) rectangle (axis cs:3.50961117554764,0.000718853321915508);
\draw[fill=color2,draw opacity=0,fill opacity=0.6] (axis cs:3.50961117554764,0) rectangle (axis cs:5.33042841830924,0.00311503106163387);
\draw[fill=color2,draw opacity=0,fill opacity=0.6] (axis cs:5.33042841830924,0) rectangle (axis cs:7.15124566107083,0.00407350215752121);
\draw[fill=color2,draw opacity=0,fill opacity=0.6] (axis cs:7.15124566107083,0) rectangle (axis cs:8.97206290383243,0.00359426660957754);
\draw[fill=color2,draw opacity=0,fill opacity=0.6] (axis cs:8.97206290383243,0) rectangle (axis cs:10.792880146594,0.00814700431504243);
\draw[fill=color2,draw opacity=0,fill opacity=0.6] (axis cs:10.792880146594,0) rectangle (axis cs:12.6136973893556,0.00670929767121141);
\draw[fill=color2,draw opacity=0,fill opacity=0.6] (axis cs:12.6136973893556,0) rectangle (axis cs:14.4345146321172,0.00958471095887343);
\draw[fill=color2,draw opacity=0,fill opacity=0.6] (axis cs:14.4345146321172,0) rectangle (axis cs:16.2553318748788,0.0143770664383102);
\draw[fill=color2,draw opacity=0,fill opacity=0.6] (axis cs:16.2553318748788,0) rectangle (axis cs:18.0761491176404,0.0184505685958314);
\draw[fill=color2,draw opacity=0,fill opacity=0.6] (axis cs:18.0761491176404,0) rectangle (axis cs:19.896966360402,0.0265975729108738);
\draw[fill=color2,draw opacity=0,fill opacity=0.6] (axis cs:19.896966360402,0) rectangle (axis cs:21.7177836031636,0.0380992260615219);
\draw[fill=color2,draw opacity=0,fill opacity=0.6] (axis cs:21.7177836031636,0) rectangle (axis cs:23.5386008459252,0.0517574391779166);
\draw[fill=color2,draw opacity=0,fill opacity=0.6] (axis cs:23.5386008459252,0) rectangle (axis cs:25.3594180886868,0.06134215013679);
\draw[fill=color2,draw opacity=0,fill opacity=0.6] (axis cs:25.3594180886868,0) rectangle (axis cs:27.1802353314484,0.0639779456504802);
\draw[fill=color2,draw opacity=0,fill opacity=0.6] (axis cs:27.1802353314484,0) rectangle (axis cs:29.00105257421,0.0486424081162827);
\draw[fill=color2,draw opacity=0,fill opacity=0.6] (axis cs:29.00105257421,0) rectangle (axis cs:30.8218698169716,0.0256391018149864);
\draw[fill=color2,draw opacity=0,fill opacity=0.6] (axis cs:30.8218698169716,0) rectangle (axis cs:32.6426870597332,0.0100639465068171);
\draw[fill=color2,draw opacity=0,fill opacity=0.6] (axis cs:32.6426870597332,0) rectangle (axis cs:34.4635043024948,0.00503197325340855);
\draw[fill=color2,draw opacity=0,fill opacity=0.6] (axis cs:34.4635043024948,0) rectangle (axis cs:36.2843215452564,0.00551120880135224);
\draw[fill=color2,draw opacity=0,fill opacity=0.6] (axis cs:36.2843215452564,0) rectangle (axis cs:38.105138788018,0.00575082657532405);
\draw[fill=color2,draw opacity=0,fill opacity=0.6] (axis cs:38.105138788018,0) rectangle (axis cs:39.9259560307796,0.00910547541092979);
\draw[fill=color2,draw opacity=0,fill opacity=0.6] (axis cs:39.9259560307796,0) rectangle (axis cs:41.7467732735412,0.0162940086300848);
\draw[fill=color2,draw opacity=0,fill opacity=0.6] (axis cs:41.7467732735412,0) rectangle (axis cs:43.5675905163028,0.027795661780733);
\draw[fill=color2,draw opacity=0,fill opacity=0.6] (axis cs:43.5675905163028,0) rectangle (axis cs:45.3884077590644,0.0277956617807329);
\draw[fill=color2,draw opacity=0,fill opacity=0.6] (axis cs:45.3884077590643,0) rectangle (axis cs:47.2092250018259,0.0335464883560571);
\draw[fill=color2,draw opacity=0,fill opacity=0.6] (axis cs:47.2092250018259,0) rectangle (axis cs:49.0300422445875,0.0162940086300848);
\draw[fill=color2,draw opacity=0,fill opacity=0.6] (axis cs:49.0300422445875,0) rectangle (axis cs:50.8508594873491,0.00575082657532408);
\draw[fill=color2,draw opacity=0,fill opacity=0.6] (axis cs:50.8508594873491,0) rectangle (axis cs:52.6716767301107,0.000479235547943671);
\addplot [ultra thick, color0]
table {%
17.8082711818504 1.73472347597681e-18
17.8082711818504 0.0836076969644542
};
\addplot [ultra thick, color1]
table {%
28.4328575134278 1.73472347597681e-18
28.4328575134278 0.0836076969644542
};
\addplot [ultra thick, color2]
table {%
26.0891374386309 1.73472347597681e-18
26.0891374386309 0.0836076969644542
};
\end{axis}

\end{tikzpicture}

%% file: figs/errors.tex
% This file was created by matplotlib2tikz v0.7.5.
\begin{tikzpicture}

\definecolor{color0}{rgb}{0.12156862745098,0.466666666666667,0.705882352941177}
\definecolor{color1}{rgb}{1,0.498039215686275,0.0549019607843137}
\definecolor{color2}{rgb}{0.172549019607843,0.627450980392157,0.172549019607843}
\definecolor{color0}{rgb}{0.83921568627451,0.152941176470588,0.156862745098039}
\definecolor{color0}{rgb}{0.6, 0.4, 0.8}
\begin{axis}[
legend cell align={left},
legend style={at={(0.03,0.97)},
nodes={scale=0.6, transform shape},
anchor=north west, draw=white!80.0!black},
tick align=outside,
tick pos=left,
x grid style={white!69.01960784313725!black},
xlabel={Future time index},
xmajorgrids,
xmin=0.1695, xmax=12.8305,
xtick style={color=black},
y grid style={white!69.01960784313725!black},
ylabel={Per-index RWSE},
ymajorgrids,
ymin=-0.02, ymax=1.0,
ytick style={color=black},
width=0.5\textwidth,
height=0.2\textwidth
]
\draw[fill=color0,draw opacity=0] (axis cs:0.745000004768372,0) rectangle (axis cs:0.915000021457672,0.0110997091978788);
\addlegendimage{ybar,ybar legend,fill=color0,draw opacity=0};
\addlegendentry{IFNN}

\draw[fill=color0,draw opacity=0] (axis cs:1.74500000476837,0) rectangle (axis cs:1.91499996185303,0.039739090949297);
\draw[fill=color0,draw opacity=0] (axis cs:2.74499988555908,0) rectangle (axis cs:2.91499996185303,0.0817820131778717);
\draw[fill=color0,draw opacity=0] (axis cs:3.74499988555908,0) rectangle (axis cs:3.91499996185303,0.145858466625214);
\draw[fill=color0,draw opacity=0] (axis cs:4.74499988555908,0) rectangle (axis cs:4.91499996185303,0.236857950687408);
\draw[fill=color0,draw opacity=0] (axis cs:5.74499988555908,0) rectangle (axis cs:5.91499996185303,0.342402338981628);
\draw[fill=color0,draw opacity=0] (axis cs:6.74499988555908,0) rectangle (axis cs:6.91499996185303,0.4459268450737);
\draw[fill=color0,draw opacity=0] (axis cs:7.74499988555908,0) rectangle (axis cs:7.91499996185303,0.531595408916473);
\draw[fill=color0,draw opacity=0] (axis cs:8.74499988555908,0) rectangle (axis cs:8.91499996185303,0.59504097700119);
\draw[fill=color0,draw opacity=0] (axis cs:9.74499988555908,0) rectangle (axis cs:9.91499996185303,0.638290643692017);
\draw[fill=color0,draw opacity=0] (axis cs:10.7449998855591,0) rectangle (axis cs:10.914999961853,0.663625717163086);
\draw[fill=color0,draw opacity=0] (axis cs:11.7449998855591,0) rectangle (axis cs:11.914999961853,0.678720116615295);
\draw[fill=color1,draw opacity=0] (axis cs:0.915000021457672,0) rectangle (axis cs:1.08499997854233,0.0199609491974115);
\addlegendimage{ybar,ybar legend,fill=color1,draw opacity=0};
\addlegendentry{CGMM}

\draw[fill=color1,draw opacity=0] (axis cs:1.91500002145767,0) rectangle (axis cs:2.08499997854233,0.0699508711695671);
\draw[fill=color1,draw opacity=0] (axis cs:2.91500002145767,0) rectangle (axis cs:3.08499997854233,0.125757202506065);
\draw[fill=color1,draw opacity=0] (axis cs:3.91500002145767,0) rectangle (axis cs:4.08499997854233,0.188439190387726);
\draw[fill=color1,draw opacity=0] (axis cs:4.91500002145767,0) rectangle (axis cs:5.08499997854233,0.27546226978302);
\draw[fill=color1,draw opacity=0] (axis cs:5.91500002145767,0) rectangle (axis cs:6.08499997854233,0.366333663463593);
\draw[fill=color1,draw opacity=0] (axis cs:6.91500002145767,0) rectangle (axis cs:7.08499997854233,0.439211130142212);
\draw[fill=color1,draw opacity=0] (axis cs:7.91500002145767,0) rectangle (axis cs:8.08499997854233,0.475391924381256);
\draw[fill=color1,draw opacity=0] (axis cs:8.91500002145767,0) rectangle (axis cs:9.08499997854233,0.479429990053177);
\draw[fill=color1,draw opacity=0] (axis cs:9.91500002145767,0) rectangle (axis cs:10.0849999785423,0.447286069393158);
\draw[fill=color1,draw opacity=0] (axis cs:10.9150000214577,0) rectangle (axis cs:11.0849999785423,0.378453224897385);
\draw[fill=color1,draw opacity=0] (axis cs:11.9150000214577,0) rectangle (axis cs:12.0849999785423,0.2989601790905);
\draw[fill=color2,draw opacity=0] (axis cs:1.08500003814697,0) rectangle (axis cs:1.25499999523163,0.0106136044487357);
\addlegendimage{ybar,ybar legend,fill=color2,draw opacity=0};
\addlegendentry{CANF}

\draw[fill=color2,draw opacity=0] (axis cs:2.08500003814697,0) rectangle (axis cs:2.25500011444092,0.0336514748632908);
\draw[fill=color2,draw opacity=0] (axis cs:3.08500003814697,0) rectangle (axis cs:3.25500011444092,0.0586150325834751);
\draw[fill=color2,draw opacity=0] (axis cs:4.08500003814697,0) rectangle (axis cs:4.25500011444092,0.0855983793735504);
\draw[fill=color2,draw opacity=0] (axis cs:5.08500003814697,0) rectangle (axis cs:5.25500011444092,0.122043162584305);
\draw[fill=color2,draw opacity=0] (axis cs:6.08500003814697,0) rectangle (axis cs:6.25500011444092,0.162660047411919);
\draw[fill=color2,draw opacity=0] (axis cs:7.08500003814697,0) rectangle (axis cs:7.25500011444092,0.188420325517654);
\draw[fill=color2,draw opacity=0] (axis cs:8.08500003814697,0) rectangle (axis cs:8.25500011444092,0.195854499936104);
\draw[fill=color2,draw opacity=0] (axis cs:9.08500003814697,0) rectangle (axis cs:9.25500011444092,0.193013489246368);
\draw[fill=color2,draw opacity=0] (axis cs:10.085000038147,0) rectangle (axis cs:10.2550001144409,0.18111315369606);
\draw[fill=color2,draw opacity=0] (axis cs:11.085000038147,0) rectangle (axis cs:11.2550001144409,0.16503943502903);
\draw[fill=color2,draw opacity=0] (axis cs:12.085000038147,0) rectangle (axis cs:12.2550001144409,0.14331391453743);
\end{axis}

\end{tikzpicture}

%% file: figs/samples.tex
% This file was created by matplotlib2tikz v0.7.5.
\begin{tikzpicture}

\definecolor{color0}{rgb}{0.12156862745098,0.466666666666667,0.705882352941177}

\begin{groupplot}[group style={group size=2 by 2}]
\nextgroupplot[
legend cell align={left},
legend style={at={(0.03,0.97)}, anchor=north west, draw=white!80.0!black,
nodes={scale=1.4, transform shape}},
tick align=outside,
tick pos=left,
x grid style={white!69.01960784313725!black},
xmin=-7.95, xmax=12.95,
xtick style={color=black},
y grid style={white!69.01960784313725!black},
ymin=0.347041374444962, ymax=2.31504618525505,
ytick style={color=black}
]
\path [draw=red, fill=red, opacity=0.2]
(axis cs:1,0.876447856426239)
--(axis cs:1,0.837636291980743)
--(axis cs:2,0.758703052997589)
--(axis cs:3,0.727728545665741)
--(axis cs:4,0.719995379447937)
--(axis cs:5,0.728747725486755)
--(axis cs:6,0.780387163162231)
--(axis cs:7,0.886351406574249)
--(axis cs:8,1.08897995948792)
--(axis cs:9,1.04104387760162)
--(axis cs:10,0.969226181507111)
--(axis cs:11,0.975309610366821)
--(axis cs:12,0.952354490756989)
--(axis cs:12,1.86252307891846)
--(axis cs:12,1.86252307891846)
--(axis cs:11,1.925173163414)
--(axis cs:10,1.94378697872162)
--(axis cs:9,1.98104393482208)
--(axis cs:8,1.95220482349396)
--(axis cs:7,1.59056508541107)
--(axis cs:6,1.15391838550568)
--(axis cs:5,0.90176522731781)
--(axis cs:4,0.836233615875244)
--(axis cs:3,0.831826984882355)
--(axis cs:2,0.823397815227509)
--(axis cs:1,0.876447856426239)
--cycle;
\addlegendimage{area legend, draw=red, fill=red, opacity=0.2}
\addlegendentry{IRNN}

\path [fill=color0, fill opacity=0.3]
(axis cs:1,0.895144820213318)
--(axis cs:1,0.819867312908173)
--(axis cs:2,0.727136135101318)
--(axis cs:3,0.682554423809052)
--(axis cs:4,0.655809760093689)
--(axis cs:5,0.669607639312744)
--(axis cs:6,0.845361471176147)
--(axis cs:7,1.12494850158691)
--(axis cs:8,1.11721909046173)
--(axis cs:9,0.981893599033356)
--(axis cs:10,1.00557708740234)
--(axis cs:11,0.967439413070679)
--(axis cs:12,0.928684592247009)
--(axis cs:12,1.11844146251678)
--(axis cs:12,1.11844146251678)
--(axis cs:11,1.17464709281921)
--(axis cs:10,1.23305225372314)
--(axis cs:9,1.24192500114441)
--(axis cs:8,1.45270466804504)
--(axis cs:7,1.50125813484192)
--(axis cs:6,1.18125379085541)
--(axis cs:5,0.897786140441895)
--(axis cs:4,0.834329724311829)
--(axis cs:3,0.845720648765564)
--(axis cs:2,0.861084759235382)
--(axis cs:1,0.895144820213318)
--cycle;
\addlegendimage{area legend, fill=color0, fill opacity=0.3}
\addlegendentry{CANF}

\addplot [semithick, black]
table {%
-7 1.91450989246368
-6 2.22559142112732
-5 2.1946325302124
-4 2.04724359512329
-3 1.88466393947601
-2 1.57605063915253
-1 1.27728247642517
0 0.981846392154694
1 0.86249828338623
2 0.824871420860291
3 0.817580342292786
4 0.822453856468201
5 0.877018392086029
6 1.11423707008362
7 1.42121744155884
8 1.37850439548492
9 1.20216774940491
10 1.22419095039368
11 1.19903516769409
12 1.14640724658966
};
\addlegendentry{True Demand}

\nextgroupplot[
tick align=outside,
tick pos=left,
x grid style={white!69.01960784313725!black},
xmin=-7.95, xmax=12.95,
xtick style={color=black},
y grid style={white!69.01960784313725!black},
ymin=0.347041374444962, ymax=2.31504618525505,
ytick style={color=black}
]
\path [draw=red, fill=red, opacity=0.2]
(axis cs:1,1.39500367641449)
--(axis cs:1,1.16690289974213)
--(axis cs:2,1.07026791572571)
--(axis cs:3,0.951948285102844)
--(axis cs:4,0.88333648443222)
--(axis cs:5,0.829918622970581)
--(axis cs:6,0.778059542179108)
--(axis cs:7,0.755945086479187)
--(axis cs:8,0.597816526889801)
--(axis cs:9,0.479221880435944)
--(axis cs:10,0.43990370631218)
--(axis cs:11,0.436496138572693)
--(axis cs:12,0.437107741832733)
--(axis cs:12,2.20210671424866)
--(axis cs:12,2.20210671424866)
--(axis cs:11,2.02006578445435)
--(axis cs:10,1.82822322845459)
--(axis cs:9,1.43199121952057)
--(axis cs:8,1.19002485275269)
--(axis cs:7,1.15415036678314)
--(axis cs:6,1.47872579097748)
--(axis cs:5,1.81060302257538)
--(axis cs:4,1.93959605693817)
--(axis cs:3,1.89612293243408)
--(axis cs:2,1.70980536937714)
--(axis cs:1,1.39500367641449)
--cycle;

\path [fill=color0, fill opacity=0.3]
(axis cs:1,1.38150143623352)
--(axis cs:1,1.1882688999176)
--(axis cs:2,1.22267782688141)
--(axis cs:3,1.21529841423035)
--(axis cs:4,1.45236229896545)
--(axis cs:5,1.38999843597412)
--(axis cs:6,1.12245643138885)
--(axis cs:7,0.877581596374512)
--(axis cs:8,0.621321022510529)
--(axis cs:9,0.578581631183624)
--(axis cs:10,0.527271568775177)
--(axis cs:11,0.501783967018127)
--(axis cs:12,0.482017517089844)
--(axis cs:12,0.738769352436066)
--(axis cs:12,0.738769352436066)
--(axis cs:11,0.734105467796326)
--(axis cs:10,0.731526434421539)
--(axis cs:9,0.757421731948853)
--(axis cs:8,0.863891780376434)
--(axis cs:7,1.18962860107422)
--(axis cs:6,1.50768327713013)
--(axis cs:5,1.85855090618134)
--(axis cs:4,1.85936498641968)
--(axis cs:3,1.80886936187744)
--(axis cs:2,1.68087959289551)
--(axis cs:1,1.38150143623352)
--cycle;

\addplot [semithick, black]
table {%
-7 0.906081736087799
-6 0.903892874717712
-5 0.885388016700745
-4 0.847667098045349
-3 0.806913256645203
-2 0.787728905677795
-1 0.823013544082642
0 0.987033247947693
1 1.32842099666595
2 1.6212123632431
3 1.81293392181396
4 1.87078964710236
5 1.72035038471222
6 1.43376207351685
7 1.15588986873627
8 0.861524045467377
9 0.755499243736267
10 0.717466950416565
11 0.714729130268097
12 0.7162064909935
};

\nextgroupplot[
tick align=outside,
tick pos=left,
x grid style={white!69.01960784313725!black},
xmin=-7.95, xmax=12.95,
xtick style={color=black},
y grid style={white!69.01960784313725!black},
ymin=0.37993596047163, ymax=1.93512176424265,
ytick style={color=black}
]
\path [draw=red, fill=red, opacity=0.2]
(axis cs:1,1.79725611209869)
--(axis cs:1,1.62580764293671)
--(axis cs:2,1.27837467193604)
--(axis cs:3,0.977562069892883)
--(axis cs:4,0.659323275089264)
--(axis cs:5,0.526979148387909)
--(axis cs:6,0.457049429416656)
--(axis cs:7,0.450626224279404)
--(axis cs:8,0.451888591051102)
--(axis cs:9,0.457221210002899)
--(axis cs:10,0.501699864864349)
--(axis cs:11,0.625514149665833)
--(axis cs:12,0.755978882312775)
--(axis cs:12,1.18106687068939)
--(axis cs:12,1.18106687068939)
--(axis cs:11,1.25496578216553)
--(axis cs:10,0.943944931030273)
--(axis cs:9,0.733140647411346)
--(axis cs:8,0.667179882526398)
--(axis cs:7,0.660284876823425)
--(axis cs:6,0.670539021492004)
--(axis cs:5,0.726784348487854)
--(axis cs:4,0.886564016342163)
--(axis cs:3,1.18852281570435)
--(axis cs:2,1.50770485401154)
--(axis cs:1,1.79725611209869)
--cycle;

\path [fill=color0, fill opacity=0.3]
(axis cs:1,1.81133103370667)
--(axis cs:1,1.64579582214355)
--(axis cs:2,1.28828465938568)
--(axis cs:3,0.980435729026794)
--(axis cs:4,0.712613821029663)
--(axis cs:5,0.593398153781891)
--(axis cs:6,0.559985458850861)
--(axis cs:7,0.559900939464569)
--(axis cs:8,0.574517905712128)
--(axis cs:9,0.61716628074646)
--(axis cs:10,0.777619540691376)
--(axis cs:11,0.956140518188477)
--(axis cs:12,0.933464288711548)
--(axis cs:12,1.19998490810394)
--(axis cs:12,1.19998490810394)
--(axis cs:11,1.29189836978912)
--(axis cs:10,1.07483267784119)
--(axis cs:9,0.90156090259552)
--(axis cs:8,0.823250651359558)
--(axis cs:7,0.819754540920258)
--(axis cs:6,0.810962200164795)
--(axis cs:5,0.852688133716583)
--(axis cs:4,0.961106181144714)
--(axis cs:3,1.23540723323822)
--(axis cs:2,1.50108122825623)
--(axis cs:1,1.81133103370667)
--cycle;

\addplot [semithick, black]
table {%
-7 0.831672489643097
-6 0.80859762430191
-5 0.84167069196701
-4 1.00487434864044
-3 1.34585571289062
-2 1.6301497220993
-1 1.81749844551086
0 1.86443150043488
1 1.69945073127747
2 1.42044603824615
3 1.15055370330811
4 0.875629603862762
5 0.767197370529175
6 0.716359794139862
7 0.723003447055817
8 0.725345253944397
9 0.787133097648621
10 0.957945942878723
11 1.17220497131348
12 1.08617305755615
};

\nextgroupplot[
tick align=outside,
tick pos=left,
x grid style={white!69.01960784313725!black},
xmin=-7.95, xmax=12.95,
xtick style={color=black},
y grid style={white!69.01960784313725!black},
ymin=0.37993596047163, ymax=1.93512176424265,
ytick style={color=black}
]
\path [draw=red, fill=red, opacity=0.2]
(axis cs:1,0.708942234516144)
--(axis cs:1,0.655189454555511)
--(axis cs:2,0.59586226940155)
--(axis cs:3,0.592447936534882)
--(axis cs:4,0.605318963527679)
--(axis cs:5,0.645559191703796)
--(axis cs:6,0.731615543365479)
--(axis cs:7,0.871800541877747)
--(axis cs:8,0.880067229270935)
--(axis cs:9,0.817365229129791)
--(axis cs:10,0.810609221458435)
--(axis cs:11,0.782774150371552)
--(axis cs:12,0.743945240974426)
--(axis cs:12,1.42366325855255)
--(axis cs:12,1.42366325855255)
--(axis cs:11,1.39408254623413)
--(axis cs:10,1.66865861415863)
--(axis cs:9,1.74437534809113)
--(axis cs:8,1.69554650783539)
--(axis cs:7,1.46049761772156)
--(axis cs:6,1.11360609531403)
--(axis cs:5,0.797670006752014)
--(axis cs:4,0.672766625881195)
--(axis cs:3,0.671275973320007)
--(axis cs:2,0.673976302146912)
--(axis cs:1,0.708942234516144)
--cycle;

\path [fill=color0, fill opacity=0.3]
(axis cs:1,0.738202214241028)
--(axis cs:1,0.644411981105804)
--(axis cs:2,0.565192103385925)
--(axis cs:3,0.526292383670807)
--(axis cs:4,0.527551412582397)
--(axis cs:5,0.567176699638367)
--(axis cs:6,0.732419312000275)
--(axis cs:7,0.899397313594818)
--(axis cs:8,0.916870474815369)
--(axis cs:9,0.815262198448181)
--(axis cs:10,0.796252489089966)
--(axis cs:11,0.74685001373291)
--(axis cs:12,0.744226217269897)
--(axis cs:12,0.947232306003571)
--(axis cs:12,0.947232306003571)
--(axis cs:11,0.957364499568939)
--(axis cs:10,1.00007772445679)
--(axis cs:9,0.990295112133026)
--(axis cs:8,1.16341078281403)
--(axis cs:7,1.22110390663147)
--(axis cs:6,0.949082493782043)
--(axis cs:5,0.713661193847656)
--(axis cs:4,0.663954138755798)
--(axis cs:3,0.670799732208252)
--(axis cs:2,0.695703029632568)
--(axis cs:1,0.738202214241028)
--cycle;

\addplot [semithick, black]
table {%
-7 1.317014336586
-6 1.60626769065857
-5 1.78717243671417
-4 1.83622336387634
-3 1.67657172679901
-2 1.37889885902405
-1 1.0919383764267
0 0.801259994506836
1 0.690346240997314
2 0.647714376449585
3 0.636920690536499
4 0.637289583683014
5 0.699279606342316
6 0.870399832725525
7 1.09750783443451
8 1.02926623821259
9 0.951966345310211
10 0.953453600406647
11 0.942996740341187
12 0.901716232299805
};
\end{groupplot}

\node at ({$(current bounding box.south west)!0.5!(current bounding box.south east)$}|-{$(current bounding box.south west)!-0.02!(current bounding box.north west)$})[
  scale=2.0,
  text=black,
  rotate=0.0
]{Time Index};
\node at ({$(current bounding box.south west)!-0.02!(current bounding box.south east)$}|-{$(current bounding box.south west)!0.5!(current bounding box.north west)$})[
  scale=2.0,
  text=black,
  rotate=90.0
]{Demand};
\end{tikzpicture}

%% file: 5-conclusion.tex
\section{Conclusion} \label{sec:discussion}

Certain real-world problems require forecasting joint multi-step uncertainties to make good decisions. Accurate multi-step short-term load probabilistic forecasting is crucial for future safe and sustainable grid operation, with applications across different hierarchies of power systems. Short-term forecasting for both generation and consumption of power is made more difficult by correlations that exist over long time horizons and noise injected by many unobserved factors.

In this paper, we used the scheduling of resource allocation under uncertainty to motivate the need to consider multi-step electric load forecasting. We extended a set of methods for joint multi-step demand forecasting and introduced the conditional approximate normalizing flow model, which estimates a joint multi-step time-series density while still being amenable to explicit conditioning. We then demonstrated the benefit of using an approximate normalizing flow for density estimation, finding the ANF to improve the KL divergence by one-third compared to a GMM when estimating the density of a uniform square. 
Finally, we showed empirically that the conditional approximate normalizing flow yields more accurate forecasts and better resource scheduling decisions when forecasting hourly electricity consumption. 
We expect CANF to work well on other low-dimensional forecasting problems with strong correlations over long time horizons. 
Future work includes addressing the co-evolution of demand forecasts at correlated locations and applying these methods to more challenging planning problems.

%% file: main.bbl
\begin{thebibliography}{}

\bibitem[\protect\BCAY{Billings}{Billings}{2013}]{billings2013narmax}
Billings, S.~A. \BBOP2013\BBCP.
\newblock {\Bem Nonlinear System Identification: {NARMAX} Methods in the Time,
  Frequency, and Spatio-Temporal Domains}.
\newblock Wiley.

\bibitem[\protect\BCAY{Blum\ \BBA\ Riedmiller}{Blum\ \BBA\
  Riedmiller}{2013}]{blum2013electricity}
Blum, M.\BBACOMMA\  \BBA\ Riedmiller, M. \BBOP2013\BBCP.
\newblock \BBOQ Electricity demand forecasting using {G}aussian processes\BBCQ\
\newblock In {\Bem Workshops at the AAAI Conference on Artificial
  Intelligence}.

\bibitem[\protect\BCAY{Bonilla, Chai,\ \BBA\ Williams}{Bonilla
  et~al.}{2008}]{bonilla2008multi}
Bonilla, E.~V., Chai, K.~M., \BBA\ Williams, C. \BBOP2008\BBCP.
\newblock \BBOQ Multi-task gaussian process prediction\BBCQ\
\newblock In {\Bem Advances in Neural Information Processing Systems
  (NeurIPS)}, \BPGS\ 153--160.

\bibitem[\protect\BCAY{Browne, Powley, Whitehouse, Lucas, Cowling, Rohlfshagen,
  Tavener, Perez, Samothrakis,\ \BBA\ Colton}{Browne
  et~al.}{2012}]{browne2012survey}
Browne, C.~B., Powley, E., Whitehouse, D., Lucas, S.~M., Cowling, P.~I.,
  Rohlfshagen, P., Tavener, S., Perez, D., Samothrakis, S., \BBA\ Colton, S.
  \BBOP2012\BBCP.
\newblock \BBOQ A survey of {M}onte {C}arlo tree search methods\BBCQ\
\newblock {\Bem IEEE Transactions on Computational Intelligence and AI in
  Games}, {\Bem 4\/}(1), 1--43.

\bibitem[\protect\BCAY{Cou{\"e}toux, Hoock, Sokolovska, Teytaud,\ \BBA\
  Bonnard}{Cou{\"e}toux et~al.}{2011}]{couetoux2011continuous}
Cou{\"e}toux, A., Hoock, J.-B., Sokolovska, N., Teytaud, O., \BBA\ Bonnard, N.
  \BBOP2011\BBCP.
\newblock \BBOQ Continuous upper confidence trees\BBCQ\
\newblock In {\Bem Learning and Intelligent Optimization (LION)}, \BPGS\
  433--445. Springer.

\bibitem[\protect\BCAY{Dinh, Sohl-Dickstein,\ \BBA\ Bengio}{Dinh
  et~al.}{2017}]{dinh2016density}
Dinh, L., Sohl-Dickstein, J., \BBA\ Bengio, S. \BBOP2017\BBCP.
\newblock \BBOQ Density estimation using real {NVP}\BBCQ\
\newblock In {\Bem International Conference on Learning Representations
  (ICLR)}.

\bibitem[\protect\BCAY{Frigola, Chen,\ \BBA\ Rasmussen}{Frigola
  et~al.}{2014}]{frigola2014variational}
Frigola, R., Chen, Y., \BBA\ Rasmussen, C.~E. \BBOP2014\BBCP.
\newblock \BBOQ Variational {G}aussian process state-space models\BBCQ\
\newblock In {\Bem Advances in Neural Information Processing Systems
  (NeurIPS)}, \BPGS\ 3680--3688.

\bibitem[\protect\BCAY{Gardner, Pleiss, Weinberger, Bindel,\ \BBA\
  Wilson}{Gardner et~al.}{2018}]{gardner2018gpytorch}
Gardner, J., Pleiss, G., Weinberger, K.~Q., Bindel, D., \BBA\ Wilson, A.~G.
  \BBOP2018\BBCP.
\newblock \BBOQ Gpytorch: Blackbox matrix-matrix {G}aussian process inference
  with {GPU} acceleration\BBCQ\
\newblock In {\Bem Advances in Neural Information Processing Systems
  (NeurIPS)}, \BPGS\ 7576--7586.

\bibitem[\protect\BCAY{Ghahramani\ \BBA\ Hinton}{Ghahramani\ \BBA\
  Hinton}{1996}]{ghahramani1996ldsem}
Ghahramani, Z.\BBACOMMA\  \BBA\ Hinton, G.~E. \BBOP1996\BBCP.
\newblock \BBOQ Parameter estimation for linear dynamical systems\BBCQ\
\newblock \BTR, CRG-TR-96-2, University of Toronto, Dept. of Computer Science.

\bibitem[\protect\BCAY{Gupta, Menda, Manchester,\ \BBA\ Kochenderfer}{Gupta
  et~al.}{2020}]{Gupta2020}
Gupta, J.~K., Menda, K., Manchester, Z., \BBA\ Kochenderfer, M.~J.
  \BBOP2020\BBCP.
\newblock \BBOQ Structured mechanical models for robot learning and
  control\BBCQ\
\newblock In {\Bem Conference on Learning for Dynamics and Control (L4DC)}.

\bibitem[\protect\BCAY{Hochreiter\ \BBA\ Schmidhuber}{Hochreiter\ \BBA\
  Schmidhuber}{1997}]{hochreiter1997long}
Hochreiter, S.\BBACOMMA\  \BBA\ Schmidhuber, J. \BBOP1997\BBCP.
\newblock \BBOQ Long short-term memory\BBCQ\
\newblock {\Bem Neural {C}omputation}, {\Bem 9\/}(8), 1735--1780.

\bibitem[\protect\BCAY{Hong\ \BBA\ Fan}{Hong\ \BBA\
  Fan}{2016}]{hong2016probabilistic}
Hong, T.\BBACOMMA\  \BBA\ Fan, S. \BBOP2016\BBCP.
\newblock \BBOQ Probabilistic electric load forecasting: A tutorial
  review\BBCQ\
\newblock {\Bem International Journal of Forecasting}, {\Bem 32\/}(3),
  914--938.

\bibitem[\protect\BCAY{Hong, Pinson,\ \BBA\ Fan}{Hong
  et~al.}{2014}]{hong2014global}
Hong, T., Pinson, P., \BBA\ Fan, S. \BBOP2014\BBCP.
\newblock \BBOQ Global energy forecasting competition 2012\BBCQ\
\newblock {\Bem International Journal of Forecasting}, {\Bem 30}, 357--363.

\bibitem[\protect\BCAY{Hong, Pinson, Fan, Zareipour, Troccoli,\ \BBA\
  Hyndman}{Hong et~al.}{2016}]{hong2016gef}
Hong, T., Pinson, P., Fan, S., Zareipour, H., Troccoli, A., \BBA\ Hyndman,
  R.~J. \BBOP2016\BBCP.
\newblock \BBOQ Probabilistic energy forecasting: Global energy forecasting
  competition 2014 and beyond\BBCQ\
\newblock {\Bem International Journal of Forecasting}, {\Bem 32}, 896--913.

\bibitem[\protect\BCAY{Hong, Xie,\ \BBA\ Black}{Hong
  et~al.}{2019}]{hong2019global}
Hong, T., Xie, J., \BBA\ Black, J. \BBOP2019\BBCP.
\newblock \BBOQ Global energy forecasting competition 2017: Hierarchical
  probabilistic load forecasting\BBCQ\
\newblock {\Bem International Journal of Forecasting}, {\Bem 35\/}(4),
  1389--1399.

\bibitem[\protect\BCAY{Hyndman\ \BBA\ Athanasopoulos}{Hyndman\ \BBA\
  Athanasopoulos}{2018}]{hyndman2018forecasting}
Hyndman, R.~J.\BBACOMMA\  \BBA\ Athanasopoulos, G. \BBOP2018\BBCP.
\newblock {\Bem Forecasting: Principles and Practice}.
\newblock OTexts.

\bibitem[\protect\BCAY{Kantas, Doucet, Singh, Maciejowski,\ \BBA\
  Chopin}{Kantas et~al.}{2015}]{kantas2015particlesurvey}
Kantas, N., Doucet, A., Singh, S.~S., Maciejowski, J., \BBA\ Chopin, N.
  \BBOP2015\BBCP.
\newblock \BBOQ On particle methods for parameter estimation in state-space
  models\BBCQ\
\newblock {\Bem Statistical Science}, {\Bem 30\/}(3), 328--351.

\bibitem[\protect\BCAY{Kingma\ \BBA\ Ba}{Kingma\ \BBA\
  Ba}{2014}]{kingma2014adam}
Kingma, D.~P.\BBACOMMA\  \BBA\ Ba, J. \BBOP2014\BBCP.
\newblock \BBOQ Adam: A method for stochastic optimization\BBCQ.

\bibitem[\protect\BCAY{Kobyzev, Prince,\ \BBA\ Brubaker}{Kobyzev
  et~al.}{2021}]{kobyzev2020normalizing}
Kobyzev, I., Prince, S.~J., \BBA\ Brubaker, M.~A. \BBOP2021\BBCP.
\newblock \BBOQ Normalizing flows: An introduction and review of current
  methods\BBCQ\
\newblock {\Bem IEEE Transactions on Pattern Analysis and Machine
  Intelligence}, {\Bem 43\/}(11), 3964--3979.

\bibitem[\protect\BCAY{Kochenderfer\ \BBA\ Wheeler}{Kochenderfer\ \BBA\
  Wheeler}{2019}]{kochenderfer2019algorithms}
Kochenderfer, M.~J.\BBACOMMA\  \BBA\ Wheeler, T.~A. \BBOP2019\BBCP.
\newblock {\Bem Algorithms for Optimization}.
\newblock MIT Press.

\bibitem[\protect\BCAY{Kong, Dong, Jia, Hill, Xu,\ \BBA\ Zhang}{Kong
  et~al.}{2017}]{kong2017short}
Kong, W., Dong, Z.~Y., Jia, Y., Hill, D.~J., Xu, Y., \BBA\ Zhang, Y.
  \BBOP2017\BBCP.
\newblock \BBOQ Short-term residential load forecasting based on {LSTM}
  recurrent neural network\BBCQ\
\newblock {\Bem IEEE Transactions on Smart Grid}, {\Bem 10\/}(1), 841--851.

\bibitem[\protect\BCAY{Markowitz}{Markowitz}{1959}]{markovitz1959portfolio}
Markowitz, H.~M. \BBOP1959\BBCP.
\newblock {\Bem Portfolio Selection: Efficient Diversification of Investments}.
\newblock Wiley.

\bibitem[\protect\BCAY{Menda, de~Becdeli{\`e}vre, Gupta, Kroo, Kochenderfer,\
  \BBA\ Manchester}{Menda et~al.}{2020}]{menda2020scalable}
Menda, K., de~Becdeli{\`e}vre, J., Gupta, J.~K., Kroo, I., Kochenderfer, M.~J.,
  \BBA\ Manchester, Z. \BBOP2020\BBCP.
\newblock \BBOQ Scalable identification of partially observed systems with
  certainty-equivalent {EM}\BBCQ\
\newblock In {\Bem International Conference on Machine Learning (ICML)}, \BPGS\
  6830--6840.

\bibitem[\protect\BCAY{Moehle, Busseti, Boyd,\ \BBA\ Wytock}{Moehle
  et~al.}{2019}]{moehle2019dynamic}
Moehle, N., Busseti, E., Boyd, S., \BBA\ Wytock, M. \BBOP2019\BBCP.
\newblock \BBOQ Dynamic energy management\BBCQ\
\newblock In {\Bem Large Scale Optimization in Supply Chains and Smart
  Manufacturing}, \BPGS\ 69--126. Springer.

\bibitem[\protect\BCAY{Murdock, Gibb, Andr{\'e}, Appavou, Brown, Epp, Kondev,
  McCrone, Musolino, Ranalder, et~al.}{Murdock
  et~al.}{2019}]{murdock2019renewables}
Murdock, H.~E., Gibb, D., Andr{\'e}, T., Appavou, F., Brown, A., Epp, B.,
  Kondev, B., McCrone, A., Musolino, E., Ranalder, L., et~al. \BBOP2019\BBCP.
\newblock \BBOQ Renewables 2019 global status report\BBCQ\
\newblock {\Bem Paris: REN21 Secretariat}, {\Bem 1}.

\bibitem[\protect\BCAY{{Nagabandi}, {Kahn}, {Fearing},\ \BBA\
  {Levine}}{{Nagabandi} et~al.}{2018}]{nagabandi2018mpc}
{Nagabandi}, A., {Kahn}, G., {Fearing}, R.~S., \BBA\ {Levine}, S.
  \BBOP2018\BBCP.
\newblock \BBOQ Neural network dynamics for model-based deep reinforcement
  learning with model-free fine-tuning\BBCQ\
\newblock In {\Bem IEEE International Conference on Robotics and Automation
  (ICRA)}, \BPGS\ 7559--7566.

\bibitem[\protect\BCAY{OPENEI}{OPENEI}{2020}]{OPENEI}
OPENEI \BBOP2020\BBCP.
\newblock \BBOQ Commercial and residential hourly load profiles for all {TMY3}
  locations in the {U}nited {S}tates\BBCQ\
\newblock \url{https://data.openei.org/submissions/153}.

\bibitem[\protect\BCAY{Owen, Maximov,\ \BBA\ Chertkov}{Owen
  et~al.}{2019}]{owen2019importance}
Owen, A.~B., Maximov, Y., \BBA\ Chertkov, M. \BBOP2019\BBCP.
\newblock \BBOQ Importance sampling the union of rare events with an
  application to power systems analysis\BBCQ\
\newblock {\Bem Electronic Journal of Statistics}, {\Bem 13\/}(1), 231--254.

\bibitem[\protect\BCAY{Pedregosa, Varoquaux, Gramfort, Michel, Thirion, Grisel,
  Blondel, Prettenhofer, Weiss, Dubourg, et~al.}{Pedregosa
  et~al.}{2011}]{scikit}
Pedregosa, F., Varoquaux, G., Gramfort, A., Michel, V., Thirion, B., Grisel,
  O., Blondel, M., Prettenhofer, P., Weiss, R., Dubourg, V., et~al.
  \BBOP2011\BBCP.
\newblock \BBOQ Scikit-learn: Machine learning in python\BBCQ\
\newblock {\Bem Journal of Machine Learning Research}, {\Bem 12}, 2825--2830.

\bibitem[\protect\BCAY{Rezende\ \BBA\ Mohamed}{Rezende\ \BBA\
  Mohamed}{2015}]{rezende2015variational}
Rezende, D.~J.\BBACOMMA\  \BBA\ Mohamed, S. \BBOP2015\BBCP.
\newblock \BBOQ Variational inference with normalizing flows\BBCQ\
\newblock In {\Bem International Conference on Machine Learning (ICML)}, \BPGS\
  1530--1538.

\bibitem[\protect\BCAY{Shepero, van~der Meer, Munkhammar,\ \BBA\
  Wid{\'e}n}{Shepero et~al.}{2018}]{shepero2018residential}
Shepero, M., van~der Meer, D., Munkhammar, J., \BBA\ Wid{\'e}n, J.
  \BBOP2018\BBCP.
\newblock \BBOQ Residential probabilistic load forecasting: A method using
  {G}aussian process designed for electric load data\BBCQ\
\newblock {\Bem Applied Energy}, {\Bem 218}, 159--172.

\bibitem[\protect\BCAY{Tribble}{Tribble}{2003}]{tribble2003relationship}
Tribble, A.~N. \BBOP2003\BBCP.
\newblock {\Bem The relationship between weather variables and electricity
  demand to improve short-term load forecasting.}
\newblock Ph.D.\ thesis, University of Oklahoma.

\bibitem[\protect\BCAY{Wang, Hertzmann,\ \BBA\ Fleet}{Wang
  et~al.}{2006}]{wang2006gpdm}
Wang, J., Hertzmann, A., \BBA\ Fleet, D.~J. \BBOP2006\BBCP.
\newblock \BBOQ Gaussian process dynamical models\BBCQ\
\newblock In {\Bem Advances in Neural Information Processing Systems
  (NeurIPS)}, \BPGS\ 1441--1448.

\bibitem[\protect\BCAY{Wang, Gan, Sun, Zhang, Lu,\ \BBA\ Kang}{Wang
  et~al.}{2019}]{wang2019probabilistic}
Wang, Y., Gan, D., Sun, M., Zhang, N., Lu, Z., \BBA\ Kang, C. \BBOP2019\BBCP.
\newblock \BBOQ Probabilistic individual load forecasting using pinball loss
  guided {LSTM}\BBCQ\
\newblock {\Bem Applied Energy}, {\Bem 235}, 10--20.

\bibitem[\protect\BCAY{Williams\ \BBA\ Rasmussen}{Williams\ \BBA\
  Rasmussen}{2006}]{williams2006gaussian}
Williams, C.~K.\BBACOMMA\  \BBA\ Rasmussen, C.~E. \BBOP2006\BBCP.
\newblock {\Bem Gaussian Processes for Machine Learning}.
\newblock MIT Press.

\bibitem[\protect\BCAY{Wu, Zeng, Lu,\ \BBA\ Boulet}{Wu
  et~al.}{2017}]{wu2017two}
Wu, D., Zeng, H., Lu, C., \BBA\ Boulet, B. \BBOP2017\BBCP.
\newblock \BBOQ Two-stage energy management for office buildings with workplace
  ev charging and renewable energy\BBCQ\
\newblock {\Bem IEEE Transactions on Transportation Electrification}, {\Bem
  3\/}(1), 225--237.

\bibitem[\protect\BCAY{Wytock, Moehle,\ \BBA\ Boyd}{Wytock
  et~al.}{2017}]{wytock2017dynamic}
Wytock, M., Moehle, N., \BBA\ Boyd, S. \BBOP2017\BBCP.
\newblock \BBOQ Dynamic energy management with scenario-based robust
  {MPC}\BBCQ\
\newblock In {\Bem American Control Conference (ACC)}, \BPGS\ 2042--2047.

\end{thebibliography}
